\pdfoutput=1

\documentclass[11pt]{article}

\usepackage{acl}

\usepackage{times}
\usepackage{latexsym}
\usepackage[normalem]{ulem}
\usepackage[T1]{fontenc}

\usepackage[utf8]{inputenc}

\usepackage{microtype}
\usepackage{graphicx}
\usepackage{xspace}
\usepackage{tabularx}
\usepackage{booktabs}
\usepackage{multirow}
\usepackage{comment}
\usepackage{caption}
\usepackage{subcaption}
\usepackage{textgreek}
\usepackage{amssymb}
\usepackage{amsmath}
\usepackage{newunicodechar}
\usepackage{color}
\usepackage{tipa}
\usepackage[T4,T1]{fontenc}
\usepackage{color, colortbl}
\usepackage{siunitx}
\usepackage{pifont} 

\newcommand*{\yoruba}{Yor\`ub\'a\xspace}
\newcommand*{\ghomala}{Ghom\'al\'a'\xspace}
\newcommand*{\ewe}{\'Ew\'e\xspace}
\newcommand*{\zulu}{isiZulu\xspace}

\newenvironment{tfour}{\fontencoding{T4}\selectfont}{}

\newcommand{\cmark}{\ding{51}}%
\newcommand{\xmark}{\ding{55}}%
\newunicodechar{ɔ}{\fon}
\newunicodechar{ɖ}{\fon}
\definecolor{Color}{gray}{0.9}

\title{A Few Thousand Translations Go A Long Way!\\ {Leveraging Pre-trained Models for African News Translation}}

% ...

\author{\normalsize David Ifeoluwa Adelani$^{1*}$, Jesujoba Oluwadara Alabi$^{2*}$, Angela Fan$^{3*}$, Julia Kreutzer$^{4*}$, \\
\textbf{\normalsize Xiaoyu Shen$^{5}$, Machel Reid$^{6*}$, Dana Ruiter$^{1}$, Dietrich Klakow$^{1}$, Peter Nabende$^{7*}$, } \\
\textbf{ \normalsize Ernie Chang$^{1*}$, Tajuddeen R. Gwadabe$^{8*}$, Freshia Sackey$^{9*}$, Bonaventure F. P. Dossou$^{10*}$, } \\
\textbf{\normalsize Chris Chinenye Emezue$^{11*}$, Colin Leong$^{12*}$, Michael Beukman$^{13*}$,} \\
\textbf{\normalsize Shamsuddeen H. Muhammad$^{14*}$, Guyo D. Jarso$^{*}$, Oreen Yousuf$^{15*}$, Andre N. Rubungo$^{16*}$,} \\
\textbf{\normalsize Gilles Hacheme$^{17*}$, Eric P. Wairagala$^{7*}$, Muhammad U. Nasir$^{18*}$, Benjamin A. Ajibade$^{*}$,} \\
\textbf{\normalsize Tunde Oluwaseyi Ajayi$^{*}$, Yvonne Wambui Gitau$^{*}$, Jade Abbott$^{*}$, Mohamed Ahmed$^{19*}$,} \\
\textbf{\normalsize Millicent Ochieng$^{19*}$, Anuoluwapo Aremu$^{*}$, Perez Ogayo$^{20*}$, Jonathan Mukiibi$^{7*}$, } \\
\textbf{\normalsize Fatoumata Ouoba Kabore$^{*}$, Godson Koffi Kalipe$^{*}$, Derguene Mbaye$^{21*}$} \\ 
\textbf{\normalsize  Allahsera Auguste Tapo$^{22*}$, Victoire M. Koagne$^{*}$, Edwin Munkoh-Buabeng$^{*}$, } \\ 
\textbf{\normalsize Valencia Wagner$^{23*}$, Idris Abdulmumin$^{24*}$, Ayodele Awokoya$^{25*}$, Happy Buzaaba$^{*}$,} \\
\textbf{\normalsize Blessing Sibanda$^{26*}$, Andiswa Bukula$^{27*}$, 
Sam Manthalu$^{28}$} \\\\
\footnotesize
$^*$Masakhane NLP, $^1$Saarland University, Germany, $^2$Inria, France, $^3$Meta AI, $^4$Google Research, $^5$Amazon Alexa AI, \\
\footnotesize
$^6$The University of Tokyo, Japan, $^{7}$Makerere University, Kampala, Uganda, $^{8}$UCAS, China, $^{9}$ JKUAT, Kenya,\\
\footnotesize
$^{10}$Jacobs University, Germany,
$^{11}$TUM, Germany,
$^{12}$University of Dayton, USA, 
$^{13}$University of the Witwatersrand, South Africa, \\
\footnotesize
$^{14}$LIAAD-INESC TEC, Porto, Portugal,
$^{15}$Uppsala University, Sweden, 
$^{16}$UPC, Spain, $^{17}$Ai4Innov $^{18}$Ominor AI\\
\footnotesize
$^{19}$Microsoft Africa Research Institute, Kenya
$^{20}$CMU, USA, $^{21}$Baamtu, $^{22}$RIT, USA, $^{23}$SPU, South Africa,\\
\footnotesize
$^{24}$ABU, Nigeria, $^{25}$UI Ibadan, Nigeria,
$^{26}$NUST, Namibia  
$^{27}$SADiLaR, South Africa,
$^{28}$University of Malawi, Malawi \\
}

\begin{document}
\maketitle
\begin{abstract}
Recent advances in the pre-training of language models leverage large-scale datasets to create multilingual models. However, low-resource languages are  mostly left out in these datasets.
This is primarily because many widely spoken languages are not well represented on the web and therefore excluded from the large-scale crawls used to create datasets. 
Furthermore, downstream users of these models are restricted to the selection of languages originally chosen for pre-training.
This work investigates how to optimally leverage existing pre-trained models to create low-resource translation systems for 16 African languages.
We focus on two questions: 1) \textit{How can pre-trained models be used for languages not included in the initial pre-training?} and 
2) \textit{How can the resulting translation models effectively transfer to new domains?}
To answer these questions, we create a 
\textit{new} African news corpus covering 16 languages, of which eight languages are not part of any existing evaluation dataset. 
We demonstrate that the most effective strategy for transferring both to additional languages and to additional domains is to fine-tune large pre-trained models on small quantities of high-quality translation data.
\end{abstract}

\section{Introduction}

Enormous efforts have been invested in making language and translation models more multilingual while leveraging the maximal amount of data for training, most prominently large crawls of monolingual and parallel data from the web~\cite{elkishky_ccaligned_2020,schwenk-etal-2021-ccmatrix,schwenk-etal-2021-wikimatrix,xue-etal-2021-MT5}.  
The resulting models are now capable of translating between hundreds of languages, including language pairs that in isolation do not have large collections of parallel data~\cite{Tang2020MultilingualTW,Xue2021ByT5TA,fan2021beyond}. 
For example, M2M-100~\citep{Goyal2021TheFE} can translate (with low accuracy) between Hausa and \yoruba, two of the most widely spoken languages in Nigeria, even though there is barely any parallel data available for training.
For languages that are not included in the set of training languages,
the model would have no knowledge on how to generate translations. Does this mean there is no hope for languages that do not have large presence on the web and are therefore not included in these pre-trained models?

We investigate \emph{how large-scale pre-trained models can be leveraged for the translation of unseen low-resource languages and domains}. We address this question by studying 16 African languages that are largely underrepresented in NLP research~\cite{joshi-etal-2020-state,nekoto_etal_2020_participatory}
and further have little to no training data available (\S\ref{sec:preliminaries}). These languages provide an ideal testbed for two challenging knowledge transfer tasks: \textbf{(1)} How can pre-trained models create translations for languages unseen at training time? and \textbf{(2)} Since training data may only exist in single domain (i.e. religious texts), 
how can a model be trained in one domain and translate another effectively at test time? 

These questions are extremely relevant for our chosen languages 
because they all have millions of native speakers and a massive need for translation technologies. For example, 
news concerning the African continent 
are almost exclusively published in English, French, or Arabic, and thereby inaccessible for speakers of only native African languages. This creates a bottleneck for information transmission, which becomes even more critical in times of crises~\citep{DBLP:journals/corr/abs-2003-11523,anastasopoulos-etal-2020-tico,DBLP:journals/corr/abs-2103-10734}.
Furthermore, the task of translating news has historically played a central role in translation research, e.g. in shared tasks since 2008~\citep{ws-2008-statistical} and as a test for determining human parity~\citep{DBLP:journals/corr/abs-1803-05567,laubli-etal-2018-machine,toral-etal-2018-attaining}. 
To spur the development of dedicated news translation models for Africa, we construct a benchmark of news translation for translating between 
16 native African languages and English or French (\S\ref{sec:corpus}). 

This allows us to compare three approaches to leveraging large-scale multilingual models for the translation of previously unseen languages: \textbf{(1)} zero-shot transfer, \textbf{(2)} continual pre-training on monolingual data, and \textbf{(3)} multi-domain fine-tuning on parallel data (\S\ref{sec:methods}). 
We find that fine-tuning pre-trained models on a few thousand sentences of high quality bitext is remarkably effective, and can be further augmented with continual pre-training on African languages and fine-tuning on news domain data (\S\ref{sec:results}).
Our contributions are the following:\footnote{All data, models and code are publicly available on \url{https://github.com/masakhane-io/lafand-mt} under academic license.} 
\begin{enumerate}
    \itemsep0em 
    \item We create a \textbf{new African news corpus} for machine translation (following principles of participatory research~\citet{nekoto_etal_2020_participatory}) covering 16 African languages.
    \item We \textbf{adapt several multilingual pre-trained models} (MT5, ByT5, mBART, M2M-100) to these largely unseen languages, and evaluate their quality on news translation.
    \item We quantify the \textbf{effectiveness of small in-domain translation sets} by measuring domain transfer effects and comparing fine-tuning strategies.
    \end{enumerate}
We find that 
having a targeted collection of translations is surprisingly effective, showcasing the power of local knowledge in so-called ``zero-resource'' scenarios~\citep{bird-2020-decolonising}.
This paints a promising picture for the development of NLP technology for understudied languages: being able to customize these models for new language of interest with as little as 2k sentences and a few fine-tuning steps, MT developers and users from any language community are less dependent on choices and monetary interest of industry powerhouses from the Global North~\citep{paulladaPower}.

\begin{table*}[th!]
 \footnotesize
 \begin{center}
 \resizebox{\textwidth}{!}{%
  \begin{tabular}{lllr|r|p{52mm}r|lr}
    \toprule
     \textbf{Target}& &\textbf{African} & \textbf{No. of}  & \textbf{Source} & \multicolumn{2}{c|}{\textbf{NEWS}}  & \multicolumn{2}{c}{\textbf{REL}} \\
    \textbf{Language} & \textbf{Family} & \textbf{Region} & \textbf{Speakers}  & \textbf{Lang.} & \textbf{Source} & \textbf{Split Sizes} & \textbf{Source} & \textbf{Total Size} \\
    \midrule
    \rowcolor{Color}
    Bambara (\texttt{bam}) & NC / Manding & West & 14M & French & Maliweb.net & 3302/ 1484/ 1600 & Bible & 28K \\
    \rowcolor{Color}
    \ghomala (\texttt{bbj}) & NC / Grassfields &Central& 1M & French & Cameroun Web & 2232/ 1133/ 1430 & Bible & 8K \\
    \rowcolor{Color}
    \ewe (\texttt{ewe}) & NC / Kwa &West& 7M & French & Benin Web TV & 2026/ 1414/ 1563 & JW300 & 618K \\
    \rowcolor{Color}
    Fon (\texttt{fon}) & NC / Volta-Niger &West& 2M & French & ORTB, Nation, Héraut, Matin Libre, LB Libéré, LE Précis, Visages.  & 2637/ 1227/ 1579 & JW300 & 32K \\
    Hausa (\texttt{hau}) & Afro-Asiatic / Chadic &West& 63M & English & WMT2021: Khamenei.v1 & 3098/ 1300/ 1500 & JW300 & 236K \\
    Igbo (\texttt{ibo}) & NC / Volta-Niger &West& 27M & English & \cite{Ezeani2020IgboEnglishMT} & 6998/ 1500/ 1500 & JW300 & 415K \\
    \rowcolor{Color}
    Luganda (\texttt{lug}) & NC / Bantu &East& 7M & English & Independent Uganda & 4075/ 1500/ 1500 & Bible & 31K \\
    \rowcolor{Color}
    Luo (\texttt{luo}) & Nilo-Saharan &East& 4M & English & Lolwe,  Standard Media & 4262/ 1500/ 1500 & Bible & 31K \\
    \rowcolor{Color}
    Mossi (\texttt{mos}) & NC / Gur &West& 8M & French & Burkina24, Lefaso & 2287/ 1478/ 1574 & JW300 & 216K \\
    \rowcolor{Color}
    Naija (\texttt{pcm}) & English-Creole &West& 75M & English & Daily Trust Nigeria & 4790/ 1484/ 1564 & JW300 & 23K \\
    Swahili (\texttt{swa}) & NC / Bantu &East \& Central & 98M & English & Global Voices, OPUS & 30782/ 1791/ 1835 & JW300 & 872K \\
    \rowcolor{Color}
    Setswana (\texttt{tsn}) & NC / Bantu &South& 14M & English & SABC News & 2100/ 1340/ 1500 & JW300 & 870K \\
    \rowcolor{Color}
    Akan/Twi (\texttt{twi}) & NC / Kwa &West& 9M & English & StarrFM, Citi News & 3337/ 1284/ 1500 & JW300 & 601K \\
    \rowcolor{Color}
    Wolof (\texttt{wol}) & NC / Senegambia &West& 5M & French & Seneweb, Jotna, Yerim Post, Socialnetlink & 3360/ 1506/ 1500 & Bible & 22K \\
    \yoruba (\texttt{yor}) & NC / Volta-Niger &West& 42M & English & \cite{adelani-etal-2021-effect} & 6644/ 1544/ 1558 & JW300 & 460K \\
    \zulu (\texttt{zul}) & NC / Bantu &South& 27M & English & \cite{rooweither_mabuya_2021_5035171}  & 3500/ 1239/ 998 & JW300 & 667K \\
    \bottomrule
  \end{tabular}
  }
  \vspace{-3mm}
  \caption{\textbf{Languages and Data Details for  MAFAND-MT Corpus}. Language, family (NC: Niger-Congo), number of speakers, news source, news (\texttt{NEWS}), and religious domain (\texttt{REL}) data split. The languages highlighted in gray did not previously have news-domain data before MAFAND-MT.}
  \vspace{-4mm}
  \label{tab:data_stat}
  \end{center}
\end{table*}

\section{Related Work}
\paragraph{African MT Datasets.}
One of the major challenges of developing MT models for African languages is lack of data. There are many attempts to automatically crawl and align sentences from the web
~\cite{schwenk-etal-2021-wikimatrix,schwenk-etal-2021-ccmatrix}. Nevertheless, the resulting corpora for many African languages are typically small and of poor quality~\cite{Kreutzer2021QualityAA}. Other cleaner parallel sources are mostly from religious sources, like the Bible covering over 1600 languages~\cite{mccarthy-etal-2020-johns} and JW300~\cite{agic-vulic-2019-jw300} from \texttt{JW.org} with over 343 languages, including over 100 African languages. Apart from the training dataset, evaluation datasets are needed to test the performance of 
multilingual MT models. The FLORES-101~\cite{Goyal2021TheFE} evaluation set, sourced from Wikipedia and manually translated, covers the largest number of languages, 
including 20 African languages. Finally, while other evaluation datasets for translating into or from African languages have been developed~\cite{Siminyu2021AI4DA,emezue-dossou-2020-ffr,Azunre2021EnglishTwiPC,Nyoni2021LowResourceNM,Gezmu2021ExtendedPC,Ali2021TowardsAP}, unfortunately there are only a few African languages with evaluation datasets in the news domain~\cite{adelani-etal-2021-effect,rooweither_mabuya_2021_5035171,Ezeani2020IgboEnglishMT} but ours covers 11 African languages (\S\ref{sec:corpus}). 

\paragraph{Low-resource MT.} 
Interest in low-resource MT has been increasing both within the MT research community~\cite{Haddow2021SurveyOL}, as well as in native speaker communities~\cite{nekoto_etal_2020_participatory,azunre2021nlp,mager-etal-2021-findings}. On the modeling side, many techniques have been developed: unsupervised MT~\cite{lample2018unsupervised} leverages monolingual data, single multilingual models capable of translating between many languages~\cite{firat-etal-2016-multi,johnson-etal-2017-googles,aharoni-etal-2019-massively,JMLR_beyond_english}, 
multilingual unsupervised models leverage a related language (with parallel data) to assist translating the low-resource language that might not even have any monolingual data~\cite{ko-etal-2021-adapting}. 
Unfortunately, unsupervised MT typically performs poorly on low-resource languages~\cite{marchisio-etal-2020-unsupervised}. 

Transfer learning from high-resource languages has achieved more promising results:
Transfer from multilingual pre-trained language models (PLM), like mBART50~\cite{Tang2020MultilingualTW} and MT5~\cite{xue-etal-2021-MT5}, and large-scale multilingual MT often outperforms bilingual MT
~\cite{tran2021facebook,Yang2021MultilingualMT}. 
For low-resource languages this strategy outperforms the baseline (Transformer) models~\cite{birch-etal-2021-surprise,adelani-etal-2021-effect,Lee2022PreTrainedMS}. The performance can be further improved by large scale pre-training~\cite{reid-etal-2021-afromt,emezue-dossou-2021-mmtafrica}.

\section{Focus Languages and Their Data}\label{sec:preliminaries}



\paragraph{Focus Languages.} We focus on 16 African languages with varying quantities of available data~\cite{joshi-etal-2020-state}, including moderately low-resource languages such as Swahili and Hausa, and very low-resource languages such as \ghomala\footnote{Spoken by an estimated 1.1M people in Cameroon} with the Bible being its largest available corpus. \autoref{tab:data_stat} provides an overview of the focus languages, including the language families, location and number of speakers, and the source and original language for our corpus. The languages are from four language families: Afro-Asiatic (e.g. Hausa), Nilo-Saharan (e.g. Luo), English Creole (e.g. Nigerian-Pidgin/Naija) and Niger-Congo. Most of the languages (13 out of 16) are from the Niger-Congo family, which is the largest language family in Africa. 
Six of the languages are predominantly spoken in Francophone countries of Africa, while the remainder are predominantly spoken in Anglophone countries of Africa.
In contrast to previous work~\citep{nekoto_etal_2020_participatory,gowda-etal-2021-many}, we do not focus exclusively on translation to/from English since this is not the primary language of the Francophone Africa community. All languages are spoken by at least one million speakers.  

\paragraph{Language Characteristics.} All languages are written in Latin script, using letters of the basic Latin alphabet
with a few omissions (e.g ``c'', ``q'', ``x'', ``z'') and additions (e.g. ``\textepsilon'', ``\textopeno'', ``\textipa{\ng}'', ``{\d o}'', including digraphs like ``gb'', ``kp'', ``gh'', and sometimes more than two-character letters). 13 of the languages are tonal, and about nine
make use of diacritics. Many African languages are morphologically rich. 
For example, all Bantu languages are agglutinative.
Fon, Mossi, and \yoruba are highly isolating.
All languages follow the Subject-Verb-Object sentence structure like English and French. ~\autoref{sec:appendix_lang_char} provides more details. 
\paragraph{Existing Parallel Corpora.}
We curate publicly available parallel data for our focus languages, which consists primarily of text in the religious domain. 
For most African languages, the largest available parallel corpora is JW300~\cite{agic-vulic-2019-jw300}, sourced from \url{jw.org},  which publishes biblical texts as well as lifestyle and opinion columns. Varying quantities of data are available for 11 of the 16 focus languages. \ewe, Igbo, Swahili, Setswana, Twi, \yoruba, and \zulu have over 400K parallel sentences.
Hausa and Mossi have slightly more than 200K parallel sentences, while Fon and Naija have around 30K sentences. For the remaining five languages that are not in the JW300 corpus,\footnote{Some languages like Luo and Luganda are covered by JW300 but are no longer available at the time of paper writing.} we make use of the Bible.\footnote{Crawled/downloaded from \url{https://ebible.org/}, except for Bambara that we obtained from \url{https://live.bible.is/} and \ghomala from \url{www.beblia.com}} We aligned the sentences automatically by the verses (around 31k in total).
\ghomala only has the New Testament with 8k
verses. Bambara and Wolof are missing some verses and books, leading to a total size of 28K and 22K.
\autoref{tab:data_stat} summarizes this information about the religious (\texttt{REL}) corpora.

\section{MAFAND-MT African News Corpus}\label{sec:corpus}





\subsection{Data Collection Process}
We introduce our newly translated news corpus; MAFAND-MT --- \underline{\textbf{M}}asakhane \underline{\textbf{A}}nglo \& \underline{\textbf{F}}ranco \underline{\textbf{A}}frica \underline{\textbf{N}}ews \underline{\textbf{D}}ataset for \underline{\textbf{M}}achine \underline{\textbf{T}}ranslation. \autoref{tab:data_stat} gives the news source and data splits for 11 African languages which includes six languages (\texttt{bam}, \texttt{bbj}, \texttt{ewe}, \texttt{fon}, \texttt{mos}, \texttt{wol}) spoken predominantly in Francophone Africa and five languages (\texttt{lug}, \texttt{luo}, \texttt{pcm}, \texttt{tsn}, \texttt{twi}) spoken predominantly in Anglophone Africa. 
The 
 MAFAND-MT corpus was created in three steps: 
\begin{enumerate} 
    \itemsep0em
\item \textbf{Crawling and preprocessing} of news websites from local newspapers that are publishing in English and French. Raw texts from the web were segmented into sentences. Most languages were crawled from one or two sites, except for Wolof and Fon that were crawled from four and seven news websites respectively due to local French language newspapers having very few articles. 
We also ensured that the articles came from a variety of topics e.g. politics, sports, culture, technology, society, religion, and education. This was carried out by native speakers of the target language with source language proficiency.
\item \textbf{Translation} of 5k--8k sentences by professional translators.
The translation process took one to four months depending on the availability of the translators.
\item \textbf{Quality control} was provided by native speakers, who discussed and, if possible, fixed problematic translations and ran automatic checks to detect misspellings, duplicated sentences, and alignment problems. 
\end{enumerate}
Following the recommendations of~\citet{nekoto_etal_2020_participatory}, we design the process to be \emph{participatory}: Everyone involved in the corpus creation is a native speaker of the respective target languages and has societal knowledge about the communities that speak those languages. This is particularly important for curation and quality control to ensure that the resulting material is appropriate and relevant for stakeholders of the final MT models~\citep{nekoto_etal_2020_participatory,Kreutzer2021QualityAA}. Furthermore, everyone received appropriate remuneration.
To enable cross-disciplinary knowledge transfer between participants in the individual steps, every language was assigned a coordinator. The coordinator conducted the initial curation in the first step, and communicated with translators and quality checkers throughout the following steps. 
\paragraph{Other Available Parallel Corpora.} We found five African languages with available parallel texts in the news domain: Hausa\footnote{https://www.statmt.org/wmt21/translation-task.html}, Igbo~\cite{Ezeani2020IgboEnglishMT}, Swahili\footnote{https://sw.globalvoices.org/}, \yoruba~\cite{adelani-etal-2021-effect}, and \zulu~\cite{rooweither_mabuya_2021_5035171}. \autoref{tab:data_stat} provides news source, the \texttt{TRAIN}, \texttt{DEV} and \texttt{TEST} splits. \autoref{sec:avail_paral_corpus} provides details on the pre-processing of the available news corpora. 
\subsection{Monolingual News Corpus}
To adapt available multilingual pre-trained models via continued pre-training to African languages, we curated texts from the 17 highest-resourced African languages and three non-native African languages that are widely spoken on the continent (Arabic, English, and French). The selection of African languages is based on their coverage in mC4~\cite{xue-etal-2021-MT5}, AfriBERTa corpora~\cite{ogueji-etal-2021-small}, and other publicly available news websites like VOA and BBC. 
We limited the size of the corpus extracted from mC4 to the first 30 million sentences (roughly 1GB of data) for Afrikaans, Amharic, Arabic, English, French, and Swahili.
In total, we collected about 12.3 GB of data. 
\autoref{sec:monolingual_corpus} provides more details about the pre-training corpus. 

\begin{table}[t]
 \begin{center}
 \scalebox{0.82}{
 \footnotesize
  \begin{tabular}{lrrp{35mm}}
    \toprule
    \textbf{Pre-trained} & \textbf{PM} & \textbf{\# African} & \\
    \textbf{Model (PM)}&  \textbf{Size} &  \textbf{Lang.} & \textbf{Focus languages covered}\\
    \midrule
    MT5/ByT5 & 580M & 13 & {hau}, {ibo}, {swa}, {yor}, {zul}  \\
    Afri[*T5] & 580M & 17 & {hau}, {ibo}, {pcm}, {swa}, {yor}, {zul}  \\
    mBART50 & 610M & 2 & {swa}  \\
    AfriMBART & 610M & 17 & {hau}, {ibo}, {pcm}, {swa}, {yor}, {zul}  \\
    M2M-100 & 418M & 17 & {hau}, {ibo}, {lug}, {swa}, {tsn}, {wol}, {yor}, {zul}  \\

    \bottomrule
  \end{tabular}
  }
  \vspace{-3mm}
  \caption{\textbf{Language coverage and size for pre-trained models}. Afri[*T5] refers to AfriMT5/ByT5.}
  \label{tab:plm_languages}
  \end{center}
  \vspace{-5mm}
\end{table}

\section{Models and Methods}~\label{sec:methods} 
\vspace{-5mm}
\subsection{Baseline Models}
We experiment with pre-trained multilingual models and our own bilingual MT baselines. We focus on pre-trained models that are approximately 500M parameters, both for computational feasibility and comparability across various different models. 
\paragraph{Transformer Baseline.} 
We train Transformer~\cite{attention_is_all_you_need} sequence-to-sequence models from scratch for each language pair using JoeyNMT~\cite{kreutzer-etal-2019-joey}. We tokenize the bitext using a joint SentencePiece\footnote{https://github.com/google/sentencepiece} unigram model~\cite{unigramSentence}, with a character coverage of 1.0 and a maximum sentence length of 4096 tokens and create a vocabulary of 10K subwords.
Models are trained on the concatenation of \texttt{REL} and \texttt{NEWS} corpora for each language.

\paragraph{Pre-trained Models.} We consider three language models, MT5~\cite{xue-etal-2021-MT5}, ByT5 (a token-free T5)~\cite{Xue2021ByT5TA}, mBART50~\cite{Tang2020MultilingualTW}, and the multilingual translation model M2M-100~\cite{fan2021beyond} for our experiments. We use MT5-base and ByT5-base, and M2M-100 with 418M parameters. \autoref{tab:plm_languages} gives the pre-trained model size, number of African languages covered, and the focus languages supported.

\begin{table*}[t]
 \footnotesize
 \begin{center}
 \resizebox{\textwidth}{!}{%
  \begin{tabular}{lrrrrrr|rrrrrrrrrr||rr}
    \toprule
     & \multicolumn{6}{c}{\textit{fr-xx}} & \multicolumn{10}{c}{\textit{en-xx}} \\
    \textbf{Model} & \textbf{bam} & \textbf{bbj} & \textbf{ewe} & \textbf{fon} & \textbf{mos} & \textbf{wol} & \textbf{hau} & \textbf{ibo} & \textbf{lug} & \textbf{luo} & \textbf{pcm} & \textbf{swa} & \textbf{tsn} & \textbf{twi} & \textbf{yor} & \textbf{zul}  & \textbf{AVG} & \textbf{MED} \\
    \midrule
    & \multicolumn{11}{c}{\textbf{BLEU}} \\
    \midrule
    M2M-100 0-shot & $-$ & $-$ & $-$ & $-$  & $-$ & $1.3$ & $0.4$   & $2.8$& $-$& $-$& $-$& $20.1$& $1.1$& $-$& $2.1$& $5.6$ & $-$ \\
    
    \midrule
    MT5 & $1.5$ & $0.4$ & $2.2$ & $1.6$ & $0.1$ & $0.9$ & $2.8$ & $18.0$ & $3.0$ & $3.1$ & $34.1$ & $25.1$ & $3.4$ & $1.7$ & $4.8$ & $11.7$ & $7.2$ & $2.9$ \\
    AfriMT5 & $2.1$ & $0.8$ & $3.7$ & $2.5$ & $0.1$ & $1.8$ & $5.1$ & $19.6$ & $5.2$ & $4.6$ & $\mathbf{35.0}$ & $\mathbf{26.7}$ & $7.0$ & $2.7$ & $6.2$ & $13.2$ & $8.5$ & $4.8$\\
    \midrule
    ByT5 & $9.5$ & $1.8$ & $5.5$ & $3.8$ & $0.1$ & $6.0$ & $8.3$ & $21.8$ & $12.1$ & $8.4$ & $30.1$ & $24.4$ & $14.7$ & $6.0$ & $7.5$ & $14.0$ & $10.9$ & $8.4$\\
    AfriByT5 & $11.4$ & $2.2$ & $5.2$ & $3.7$ & $0.2$ & $6.4$ & $9.3$ & $22.7$ & $13.1$ & $8.9$ & $30.0$ & $24.7$ & $17.0$ & $6.1$ & $7.6$ & $15.3$ & $11.5$ & $9.1$\\
    \midrule
    mBART50 & $18.6$ & $2.4$ & $5.3$ & $6.2$ & $0.8$ & $9.7$ & $8.9$ & $21.1$ & $12.0$ & $10.0$ & $34.1$ & $25.8$ & $16.8$ & $7.5$ & $10.0$ & $\mathbf{21.2}$ & $13.2$ & $10.0$\\
    AfriMBART & $15.3$ & $2.4$ & $5.7$ & $4.4$ & $0.6$ & $8.6$ & $10.4$ & $22.4$ & $10.0$ & $9.8$ & $30.0$ & $22.7$ & $12.8$ & $6.3$ & $9.6$ & $20.1$ & $11.9$ & $9.9$\\
    \midrule
    M2M-100 & $\mathbf{22.7}$ & $\mathbf{2.9}$ & $\mathbf{6.4}$ & $\mathbf{7.1}$ & $\mathbf{1.0}$ & $\mathbf{12.4}$ & $\mathbf{16.0}$ & $\mathbf{24.7}$ & $\mathbf{14.3}$ & $\mathbf{11.5}$ & $33.9$ & $\mathbf{26.7}$ & $\mathbf{24.7}$ & $\mathbf{8.8}$ & $\mathbf{12.8}$ & $21.0$ & $\mathbf{15.4}$ & $\mathbf{13.6}$\\
    M2M-100-EN/FR & $18.5$ & $2.2$ & $6.2$ & $4.3$ & $0.8$ & $10.6$ & $7.0$ & $22.4$ & $8.9$ & $9.5$ & $34.9$ & $26.4$ & $19.7$ & $7.0$ & $5.6$ & $15.6$ & $12.5$ & $9.2$ \\
    \bottomrule
    \midrule
    & \multicolumn{11}{c}{\textbf{CHRF}} \\
    \midrule
    M2M-100 0-shot & $-$ & $-$ & $-$ & $-$  & $-$ & $4.3$ & $12.4$   & $19.0$& $-$& $-$& $-$& $47.7$& $8.7$& $-$& $10.4$& $20.1$ & $-$ \\
    
    \midrule
    MT5 & $10.0$ & $7.4$ & $9.7$ & $11.5$ & $7.9$ & $9.1$ & $23.6$ & $41.1$ & $24.9$ & $21.6$ & $64.1$ & $53.7$ & $22.8$ & $17.8$ & $20.8$ & $36.0$ & $23.9$ & $21.2$\\

    AfriMT5 & $14.0$ & $12.7$ & $16.6$ & $14.8$ & $8.2$ & $13.8$ & $29.7$ & $43.1$ & $30.4$ & $25.7$ & $\mathbf{64.7}$ & $55.1$ & $31.5$ & $21.5$ & $24.3$ & $40.3$ & $27.9$ & $25.0$\\
    \midrule
    ByT5 & $27.8$ & $17.7$ & $23.8$ & $16.1$ & $8.8$ & $22.9$ & $31.3$ & $46.5$ & $40.0$ & $32.2$ & $58.1$ & $52.5$ & $38.6$ & $27.9$ & $25.5$ & $40.3$ & $31.9$ & $29.6$\\
    AfriByT5 & $31.4$ & $19.9$ & $24.1$ & $16.5$ & $9.8$ & $23.8$ & $32.8$ & $47.4$ & $42.2$ & $33.6$ & $58.0$ & $52.8$ & $42.1$ & $29.0$ & $26.0$ & $42.9$ & $33.3$ & $32.1$\\
    \midrule
    mBART50 & $42.3$ & $22.0$ & $27.7$ & $25.7$ & $16.0$ & $31.9$ & $32.6$ & $45.9$ & $41.1$ & $36.7$ & $64.2$ & $54.4$ & $43.0$ & $35.6$ & $31.1$ & $50.2$ & $37.5$ & $36.2$\\
    AfriMBART & $40.4$ & $20.1$ & $26.9$ & $24.1$ & $15.1$ & $30.9$ & $40.3$ & $47.4$ & $38.6$ & $36.7$ & $54.9$ & $52.7$ & $40.3$ & $34.2$ & $31.1$ & $49.3$ & $36.4$ & $37.7$\\
    \midrule
    M2M-100 & $\mathbf{48.2}$ & $\mathbf{23.1}$ & $\mathbf{30.9}$ & $\mathbf{27.6}$ & $\mathbf{16.7}$ & $\mathbf{35.7}$ & $\mathbf{43.3}$ & $\mathbf{50.0}$ & $\mathbf{45.5}$ & $\mathbf{39.0}$ & $64.0$ & $\mathbf{56.4}$ & $\mathbf{52.0}$ & $\mathbf{38.2}$ & $\mathbf{35.9}$ & $\mathbf{51.2}$ & $\mathbf{41.1}$ & $\mathbf{41.2}$\\
    M2M-100-EN/FR & $43.4$ & $20.6$ & $29.4$ & $23.2$ & $16.3$ & $32.8$ & $33.3$ & $46.9$ & $38.8$ & $36.5$ & $64.5$ & $55.4$ & $47.1$ & $33.6$ & $25.3$ & $42.9$ & $36.9$ & $35.0$ \\
    \bottomrule
    
  \end{tabular}
  }
  \vspace{-3mm}
  \caption{\textbf{Results adding African Languages to Pre-Trained Models, en/fr-xx}. We calculate BLEU and CHRF on the news domain when training on only \texttt{NEWS} data from MAFAND-MT.}
  \vspace{-4mm}
  \label{tab:news_fr_en_xx}
  \end{center}
\end{table*}

\begin{table*}[t]
 \footnotesize
 \begin{center}
 \resizebox{\textwidth}{!}{%
  \begin{tabular}{lrrrrrr|rrrrrrrrrr||rr}
    \toprule
     & \multicolumn{6}{c}{\textit{xx-fr}} & \multicolumn{10}{c}{\textit{xx-en}} \\
    \textbf{Model} & \textbf{bam} & \textbf{bbj} & \textbf{ewe} & \textbf{fon} & \textbf{mos} & \textbf{wol} & \textbf{hau} & \textbf{ibo} & \textbf{lug} & \textbf{luo} & \textbf{pcm} & \textbf{swa} & \textbf{tsn} & \textbf{twi} & \textbf{yor} & \textbf{zul}  & \textbf{AVG} & \textbf{MED} \\
    \midrule
    \midrule
    & \multicolumn{11}{c}{\textbf{BLEU}} \\
    \midrule
    M2M-100 0-shot & $-$ & $-$ & $-$ & $-$  & $-$ & $0.8$ & $2.2$   & $6.4$& $-$& $-$& $-$& $25.2$& $3.3$& $-$& $3.0$& $13.8$ & $-$ \\
    \midrule
    MT5 & $2.5$ & $0.9$ & $1.1$ & $2.4$ & $0.7$ & $1.3$ & $5.8$ & $18.9$ & $12.6$ & $6.4$ & $42.2$ & $29.5$ & $9.5$ & $4.6$ & $12.3$ & $22.4$ & $10.8$ & $6.1$ \\
    AfriMT5 & $6.4$ & $2.0$ & $2.1$ & $4.2$ & $1.2$ & $2.9$ & $10.4$ & $19.5$ & $15.5$ & $9.7$ & $44.6$ & $30.6$ & $16.1$ & $8.4$ & $13.8$ & $24.0$ & $13.2$ & $10.0$ \\
    \midrule
    ByT5 & $10.0$ & $2.7$ & $4.1$ & $4.9$ & $1.5$ & $7.2$ & $12.9$ & $21.0$ & $19.8$ & $12.1$ & $39.4$ & $27.1$ & $18.6$ & $9.8$ & $11.5$ & $22.8$ & $14.1$ & $11.8$ \\
    AfriByT5 & $13.8$ & $4.4$ & $4.5$ & $5.8$ & $2.2$ & $9.0$ & $13.5$ & $20.7$ & $\mathbf{21.1}$ & $12.5$ & $39.5$ & $27.0$ & $19.7$ & $10.5$ & $11.9$ & $24.0$ & $15.0$ & $13.0$ \\
    \midrule
    mBART50 & $6.8$ & $0.3$ & $1.7$ & $0.8$ & $0.6$ & $6.3$ & $11.5$ & $13.2$ & $14.5$ & $9.1$ & $44.2$ & $29.0$ & $2.0$ & $0.5$ & $8.1$ & $31.1$ & $11.2$ & $7.4$ \\
    AfriMBART & $8.1$ & $2.3$ & $3.0$ & $4.5$ & $1.7$ & $3.2$ & $10.2$ & $15.5$ & $13.1$ & $8.0$ & $43.7$ & $29.2$ & $7.2$ & $6.5$ & $9.5$ & $33.0$ & $12.4$ & $8.0$ \\
    \midrule
    M2M-100 & $\mathbf{22.1}$ & $\mathbf{5.4}$ & $6.9$ & $8.4$ & $\mathbf{2.8}$ & $10.3$ & $\mathbf{17.0}$ & $19.0$ & $20.0$ & $13.0$ & $43.8$ & $29.8$ & $20.0$ & $10.9$ & $\mathbf{16.0}$ & $\mathbf{37.8}$ & $\mathbf{17.7}$ & $\mathbf{16.5}$ \\
    M2M-100-EN/FR & $\mathbf{22.1}$ & $5.1$ & $\mathbf{7.4}$ & $\mathbf{9.1}$ & $2.1$ & $\mathbf{10.5}$ & $11.4$ & $20.3$ & $19.8$ & $\mathbf{14.0}$ & $\mathbf{45.2}$ & $\mathbf{30.0}$ & $\mathbf{21.4}$ & $\mathbf{11.7}$ & $13.4$ & $9.5$ & $15.8$ & $12.6$ \\
    \bottomrule
    \midrule
    & \multicolumn{11}{c}{\textbf{CHRF}} \\
    \midrule
    M2M-100 0-shot & $-$ & $-$ & $-$ & $-$  & $-$ & $12.3$ & $23.7$   & $29.7$& $-$& $-$& $-$& $51.6$& $21.1$& $-$& $18.3$& $35.7$ & $-$ \\
    \midrule
    MT5 & $19.4$ & $15.1$ & $17.0$ & $17.9$ & $10.9$ & $16.2$ & $26.3$ & $43.5$ & $36.3$ & $26.1$ & $66.9$ & $53.7$ & $32.2$ & $25.2$ & $31.1$ & $43.9$ & $30.1$ & $26.2$ \\
    AfriMT5 & $27.7$ & $19.6$ & $21.1$ & $21.4$ & $13.2$ & $21.6$ & $32.5$ & $44.9$ & $40.2$ & $32.2$ & $68.4$ & $54.5$ & $39.6$ & $31.2$ & $33.9$ & $45.9$ & $34.2$ & $32.4$ \\
    \midrule
    ByT5 & $31.2$ & $21.8$ & $24.8$ & $20.5$ & $15.4$ & $26.2$ & $33.2$ & $46.4$ & $45.4$ & $34.1$ & $62.0$ & $50.6$ & $42.4$ & $32.9$ & $31.4$ & $42.5$ & $35.0$ & $33.0$ \\
    AfriByT5 & $34.8$ & $25.5$ & $24.9$ & $22.0$ & $16.2$ & $29.3$ & $33.9$ & $46.4$ & $\mathbf{47.1}$ & $35.0$ & $62.1$ & $50.5$ & $43.4$ & $33.4$ & $32.0$ & $43.7$ & $36.3$ & $34.3$ \\
    \midrule
    mBART50 & $26.0$ & $17.1$ & $20.9$ & $20.2$ & $17.1$ & $26.6$ & $32.0$ & $37.9$ & $39.0$ & $31.0$ & $68.2$ & $53.5$ & $20.1$ & $19.4$ & $26.7$ & $49.0$ & $31.5$ & $26.6$ \\
    AfriMBART & $31.4$ & $22.9$ & $27.2$ & $26.3$ & $17.0$ & $25.0$ & $34.3$ & $42.0$ & $40.4$ & $29.8$ & $67.8$ & $53.5$ & $31.4$ & $30.6$ & $30.0$ & $51.7$ & $35.1$ & $31.0$ \\
    \midrule
    M2M-100 & $\mathbf{45.9}$ & $26.5$ & $30.9$ & $27.5$ & $\mathbf{17.7}$ & $33.8$ & $\mathbf{38.7}$ & $46.1$ & $46.4$ & $36.7$ & $68.6$ & $54.8$ & $45.2$ & $35.1$ & $\mathbf{38.1}$ & $\mathbf{55.5}$ & $\mathbf{40.5}$ & $\mathbf{38.4}$ \\
    M2M-100-EN/FR & $45.6$ & $\mathbf{26.9}$ & $\mathbf{32.2}$ & $\mathbf{28.7}$ & $17.0$ & $\mathbf{34.3}$ & $35.1$ & $\mathbf{46.6}$ & $46.0$ & $\mathbf{37.6}$ & $\mathbf{69.0}$ & $\mathbf{55.0}$ & $\mathbf{46.3}$ & $\mathbf{36.0}$ & $35.2$ & $31.5$ & $38.9$ & $35.6$ \\
    \bottomrule
  \end{tabular}
  }
  \vspace{-2mm}
  \caption{\textbf{Results adding African Languages to Pre-Trained Models, xx-en/fr}. We calculate BLEU and CHRF on the news domain when training on only \texttt{NEWS} data from MAFAND-MT.}
  \label{tab:news_xx_fr_en}
  \end{center}
\end{table*}

\vspace{2mm}
\subsection{Transfer Learning Across Languages}

We describe two methods for adding new languages to existing models: continual pre-training and many-to-many multilingual translation.

\paragraph{Continual Pre-training.}
The effectiveness of PLMs is limited on extremely low-resource languages because they rarely, if ever, occur in the pre-training corpus~\cite{wang-etal-2020-extending,liu-etal-2021-continual}. As shown in Table~\ref{tab:plm_languages}, even for MT5 and M2M-100, which cover 100 languages, less than half of the African languages under study are included. To adapt the existing PLMs to our languages corpora and domains, we apply continual pre-training~\cite{gururangan2020don,liu-etal-2021-continual} using our collected monolingual corpus. Specifically, before fine-tuning on the parallel MT data, models are pre-trained with their original training objective and vocabulary\footnote{Changing the vocabulary~\cite{gururangan2020don} to fit the languages, or adding MT-focused training objectives for word alignment~\cite{liu-etal-2021-continual} can potentially improve the performance further, which we leave for future work.} on the monolingual corpus. Pre-training parameters can be found in the appendix. We refer to the models  adapted to African languages as AfriMT5, AfriByT5, and AfriMBART. 

\paragraph{Many-to-Many Translation.}

We fine-tuned M2M-100 for African multilingual translation to create English- and French-centric models. For the English-centric model, the M2M-100 model was fine-tuned on the news data for \texttt{en}--\{\texttt{hau}, \texttt{ibo}, \texttt{lug}, \texttt{luo}, \texttt{pcm}, \texttt{swa}, \texttt{tsn}, \texttt{twi}, \texttt{yor}, \texttt{zul}\} while the French-centric model is trained on \texttt{fr}--\{\texttt{bam}, \texttt{bbj}, \texttt{ewe},  \texttt{fon}, \texttt{mos}, \texttt{wol}\}. 
Languages not included in the pre-trained M2M-100 model were assigned the language code of a language included in M2M-100 but excluded from our study.

\subsection{Transfer Learning Across Domains}
As there is very limited MT data on the news domain, we compare different methods that combine the \textit{large} data from the religious domain (\texttt{REL}) and the \textit{small} data from the NEWS domain (\texttt{NEWS}) to fine-tune M2M-100:
\vspace{-3mm}
\begin{enumerate}
    \itemsep0em
    \item \texttt{REL+NEWS}: Fine-tuning on the aggregation of \texttt{REL} and \texttt{NEWS}. 
    \item \texttt{REL$\rightarrow$NEWS}: Training on \texttt{REL}, followed by fine-tuning on \texttt{NEWS}. 
    \item \texttt{REL+NEWS$\rightarrow$NEWS}: \texttt{REL+NEWS}, followed by additional fine-tuning on \texttt{NEWS}. 
    \vspace{-2mm}
\end{enumerate}

Each fine-tuning stage lasts for three epochs.
We evaluate translation quality with BLEU~\cite{papineni-etal-2002-bleu} using SacreBLEU~\citep{post-2018-call}\footnote{``intl'' tokenizer, all data comes untokenized.} and ChrF~\cite{popovic-2015-chrf}. 

\section{Results and Discussion}\label{sec:results}

We successfully adapt several multilingual pre-trained models to previously unseen African languages and quantify the effectiveness of small in-domain translation datasets. 
We discuss the effects of domain shift and analyze mitigation strategies.

\subsection{Adaptation to the Focus Languages}
\label{sec:adapt_focus_lang}
We demonstrate that fine-tuning with a few thousand high-quality bitext is effective for adding new languages to pre-trained models. Further, continuing to pre-train to specialize models to African languages 
further improves performance. 

\paragraph{Zero-Shot Translation.} 
\autoref{tab:news_fr_en_xx} and \autoref{tab:news_xx_fr_en} gives the result of zero-shot evaluation on \texttt{NEWS}. 
We evaluate only on the M2M-100 dataset because it has been pre-trained on parallel texts with a few of our focus languages.
We observe very poor performance ($<5$ BLEU) on the languages except for \texttt{zul} ($>13$ BLEU) and \texttt{swa} ($>20$ BLEU)  in both translation directions. 
For \texttt{swa}, its likely that the performance is reasonable because M2M-100 has seen more bitext during pre-training ($2.4$M sentences in CCAligned~\cite{elkishky_ccaligned_2020}). 
Other African languages except for Afrikaans have less than 600K sentences in CCAligned, and are also of a lower quality~\citep{Kreutzer2021QualityAA} which affect  overall zero-shot performance. 

\paragraph{Performance after Fine-tuning.} 
We found impressive performance after fine-tuning PLMs and M2M-100 on few thousand sentences (mostly 2K--7K sentences, except for \texttt{swa} with 30K sentences), including languages not seen during pre-training. 
For \textit{en/fr-xx}, MT5 has a poor transfer performance with average BLEU of $7.2$, despite being pre-trained on 101 languages. 
ByT5 outperforms MT5 by over $3$ BLEU on average, even though their performances were reported to be similar in previous work~\citep{Xue2021ByT5TA}. 
This indicates that ByT5 might be preferable over MT5 when translating low-resource languages. 
Surprisingly, mBART50 that was only pre-trained on 50 languages and 2 African languages outperformed MT5 and ByT5 which are pre-trained on 101 languages. Overall, we found M2M-100 to be the best model, most likely because it was pre-trained on a translation task. 
In general, BLEU scores are relatively low
($<15$ BLEU for 9 out of 16 languages for en/fr-xx and 7 in xx-en/fr) even when fine-tuning M2M-100 on in-domain data, which suggests that developing more effective methods for fine-tuning might be a promising future direction. 
The languages with the best quality according to BLEU on the target side are \texttt{pcm}, \texttt{swa} and \texttt{tsn},  
and
\texttt{pcm}, \texttt{zul}, and \texttt{swa} on the source side.

BLEU scores are 
higher when translating from an African language, which is expected due to the more frequent exposure to English and French on the target side during pre-training, and BLEU being penalized more for morphologically rich languages like \texttt{bbj}, \texttt{lug}, \texttt{swa}, \texttt{tsn}, and \texttt{zul}). The ChrF metric works better for them.
For example, fine-tuning M2M-100 on \texttt{NEWS} and evaluating on \texttt{zul} has a BLEU of $21.0$ in \textit{en/fr-xx}, and BLEU of $37.8$ in the \textit{xx-en/fr} showing a large gap in performance in both directions. However, with the ChrF, we find a smaller performance gap 
($51.2$ in \textit{en/fr-xx} and $55.5$ in the \textit{xx-en/fr}.

\paragraph{Continual Pre-training.} We observe an improvement in BLEU when we utilize AfriMT5 and AfriByT5, for languages included in our continual pre-training corpus (\autoref{sec:monolingual_corpus}). Other languages also benefit despite not being seen during continual pre-training, possibly due to language similarity. 
For example, AfriByT5 on \textit{fr-bam} improved by $1.9$ BLEU over ByT5 and AfriMT5 on \textit{en-tsn} improved by $3.6$ BLEU over MT5. 
On average, AfriMT5 improved over MT5 by $1.3$ BLEU in \texttt{en/fr-xx} and $2.4$ BLEU in the \textit{xx-en/fr}. 
The improvement for AfriByT5 was much smaller: $0.6$ and $0.9$ BLEU in \textit{en/fr-xx} and  \textit{xx-en/fr} translation directions. 
For AfriMBART, we did not see any improvement on average, only the performance of \texttt{hau} ($1.5$ BLEU) and \texttt{ibo} ($0.7$ BLEU) improved in \textit{en/fr-xx} direction. 
However, in the \textit{xx-en/fr} direction, \texttt{fon}, \texttt{tsn}, \texttt{twi}, and \texttt{zul} improved by 2.7--6.0 BLEU. 

\paragraph{Many-to-Many Multilingual MT.}  Training on the combined news corpus from all languages that use French or English separately does not appear to help much. We see slight improvements for most languages only in the \textit{xx-en/fr} direction.

\subsection{Adaptation to the News Domain}

\begin{table*}[t]
 \footnotesize
 \begin{center}
 \resizebox{\textwidth}{!}{%
  \begin{tabular}{lrrrrrr|rrrrrrrrrr||rr}
    \toprule
     & \multicolumn{6}{c}{\textit{fr-xx}} & \multicolumn{10}{c}{\textit{en-xx}} \\
    \textbf{Model} & \textbf{bam} & \textbf{bbj} & \textbf{ewe} & \textbf{fon} & \textbf{mos} & \textbf{wol} & \textbf{hau} & \textbf{ibo} & \textbf{lug} & \textbf{luo} & \textbf{pcm} & \textbf{swa} & \textbf{tsn} & \textbf{twi} & \textbf{yor} & \textbf{zul}  & \textbf{AVG} & \textbf{MED} \\
    \midrule
    & \multicolumn{11}{c}{\textbf{BLEU}} \\
    \midrule
    \multicolumn{2}{l}{Transformer} \\
    \texttt{REL+NEWS} & $7.3$ & $0.1$ & $6.2$ & $2.9$ & $2.1$ & $3.1$ & $10.7$ & $22.4$ & $4.6$ & $3.7$ & $11.7$ & $26.2$ & $28.1$ & $8.7$ & $9.7$ & $16.5$ & $10.2$ & $8.0$ \\
    \texttt{REL$\rightarrow$NEWS} & $5.1$ & $0.2$ & $5.4$ & $2.8$ & $1.7$ & $2.3$ & $11.7$ & $22.7$ & $3.9$ & $3.3$ & $11.9$ & $26.3$ & $29.7$ & $8.7$ & $8.4$ & $20.3$ & $10.3$ & $6.9$ \\
    \texttt{REL+NEWS$\rightarrow$NEWS} & $8.5$ & $0.3$ & $6.5$ & $3.2$ & $\mathbf{2.2}$ & $3.7$ & $12.0$ & $23.6$ & $5.1$ & $4.3$ & $13.8$ & $26.6$ & $29.3$ & $9.0$ & $9.7$ & $20.1$ & $11.1$ & $8.8$ \\
    \midrule
    \multicolumn{2}{l}{M2M-100} \\
    \texttt{REL+NEWS} & $23.0$ & $2.8$ & $7.7$ & $6.5$ & $0.9$ & $11.2$ & $12.9$ & $24.7$ & $13.9$ & $11.6$ & $\mathbf{35.1}$ & $23.3$ & $29.0$ & $9.7$ & $12.4$ & $18.3$ & $15.2$ & $12.6$ \\
    \texttt{REL$\rightarrow$NEWS} & $20.3$ & $\mathbf{3.1}$ & $7.7$ & $\mathbf{7.5}$ & $1.1$ & $12.0$ & $15.0$ & $\mathbf{26.0}$ & $15.4$ & $11.9$ & $35.0$ & $\mathbf{27.7}$ & $\mathbf{31.9}$ & $10.0$ & $13.4$ & $\mathbf{22.9}$ & $16.3$ & $14.2$ \\
    \texttt{REL+NEWS$\rightarrow$NEWS} & $\mathbf{24.7}$ & $\mathbf{3.1}$ & $\mathbf{8.9}$ & $7.4$ & $1.1$ & $\mathbf{12.7}$ & $\mathbf{15.9}$ & $25.8$ & $\mathbf{15.7}$ & $\mathbf{12.0}$ & $34.2$ & $27.3$ & $\mathbf{31.9}$ & $\mathbf{10.2}$ & $\mathbf{13.9}$ & $22.6$ & $\mathbf{16.7}$ & $\mathbf{14.8}$  \\
    
    \bottomrule 
    \toprule 
    & \multicolumn{11}{c}{\textbf{CHRF}} \\
    \midrule
    \multicolumn{2}{l}{Transformer} \\
    \texttt{REL+NEWS} & $25.6$ & $9.6$ & $30.6$ & $14.5$ & $17.7$ & $18.9$ & $36.7$ & $46.7$ & $30.5$ & $26.4$ & $37.8$ & $55.3$ & $55.0$ & $36.7$ & $30.6$ & $50.0$ & $32.7$ & $30.6$ \\
    \texttt{REL$\rightarrow$NEWS} & $18.2$ & $11.2$ & $27.1$ & $15.4$ & $18.3$ & $15.9$ & $37.4$ & $47.2$ & $28.7$ & $24.4$ & $38.3$ & $55.5$ & $56.3$ & $36.6$ & $28.9$ & $53.0$ & $32.0$ & $28.8$ \\
    \texttt{REL+NEWS$\rightarrow$NEWS} & $27.4$ & $12.8$ & $31.5$ & $16.5$ & $19.9$ & $20.2$ & $38.3$ & $48.3$ & $30.6$ & $27.7$ & $42.6$ & $55.6$ & $56.3$ & $37.7$ & $30.6$ & $53.4$ & $34.3$ & $31.0$ \\
    \midrule
    \multicolumn{2}{l}{M2M-100} \\
    \texttt{REL+NEWS} & $46.8$ & $22.1$ & $36.7$ & $26.2$ & $16.0$ & $33.5$ & $38.4$ & $50.1$ & $44.5$ & $38.1$ & $64.7$ & $53.0$ & $57.2$ & $39.7$ & $35.2$ & $53.1$ & $41.0$ & $39.0$ \\
    \texttt{REL$\rightarrow$NEWS} & $44.1$ & $22.6$ & $34.1$ & $27.7$ & $16.8$ & $34.7$ & $41.3$ & $51.3$ & $45.6$ & $38.6$ & $\mathbf{64.7}$ & $\mathbf{57.2}$ & $59.3$ & $40.6$ & $37.1$ & $\mathbf{56.3}$ & $42.0$ & $41.0$ \\
    \texttt{REL+NEWS$\rightarrow$NEWS} & $\mathbf{49.9}$ & $\mathbf{23.5}$ & $\mathbf{37.5}$ & $\mathbf{28.5}$ & $\mathbf{16.8}$ & $\mathbf{35.8}$ & $\mathbf{42.1}$ & $\mathbf{51.3}$ & $\mathbf{46.9}$ & $\mathbf{39.4}$ & $64.2$ & $57.0$ & $\mathbf{59.5}$ & $\mathbf{40.8}$ & $\mathbf{37.4}$ & $\mathbf{56.3}$ & $\mathbf{42.9}$ & $\mathbf{41.4}$ \\

    \bottomrule
  \end{tabular}
  }
  \vspace{-1mm}
  \caption{\textbf{Results adapting to Domain Shift, en/fr-xx}. We calculate BLEU and ChrF on the news domain when training on different combinations of \texttt{REL} and \texttt{NEWS}.}
  \label{tab:rel_news_fr_en_xx}
  \end{center}
\end{table*}

\begin{figure*}[t]
\centering
    \hspace{-2.5\baselineskip}
    \begin{subfigure}{0.47\textwidth}
        \centering
        \includegraphics[width=0.8\textwidth]{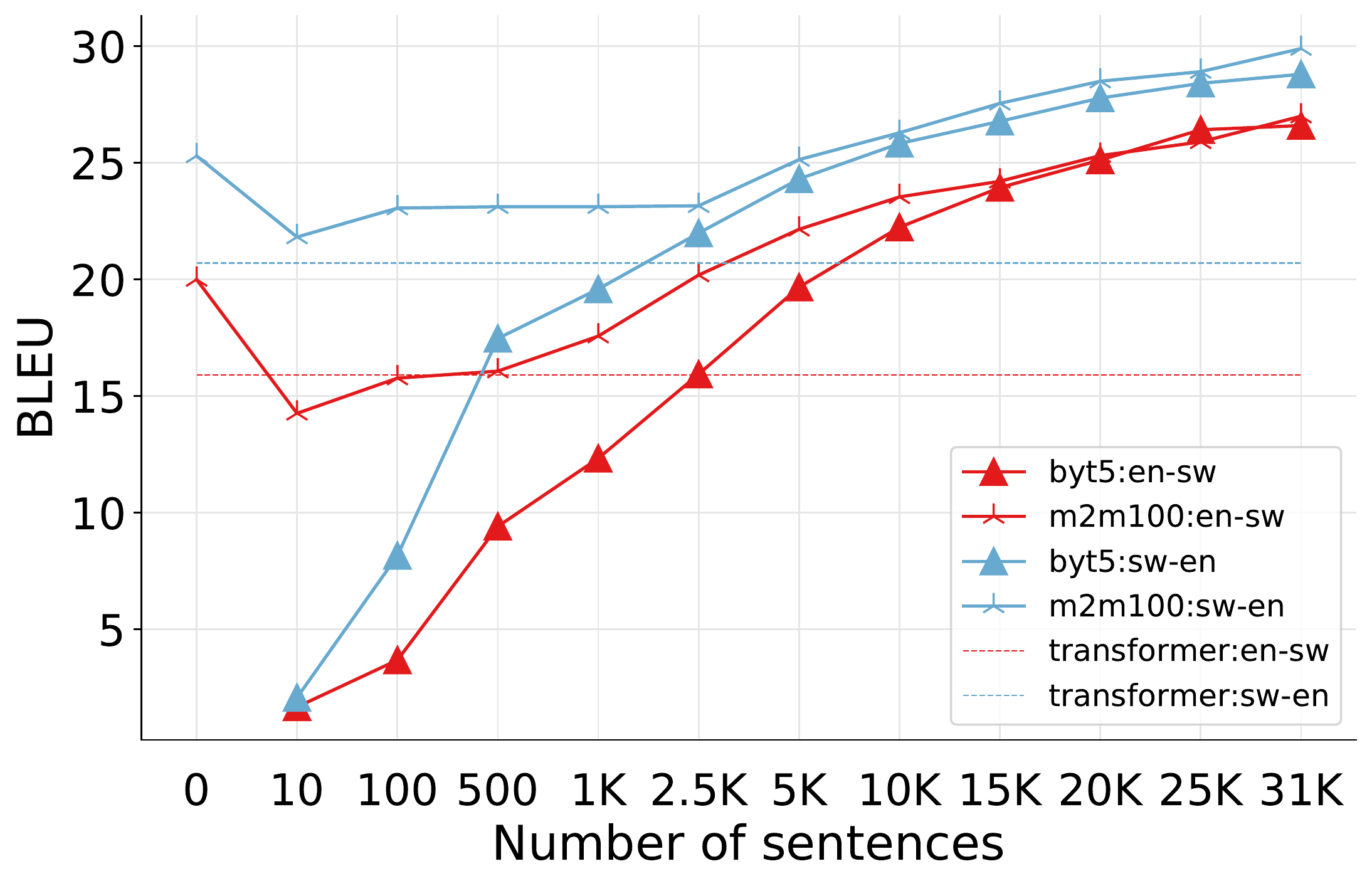}
        \caption{eng-swa} \label{fig:sw_n}
    \end{subfigure}
    \hspace{-5.5\baselineskip}
    \begin{subfigure}{0.45\textwidth}
        \centering
        \includegraphics[width=0.6\textwidth]{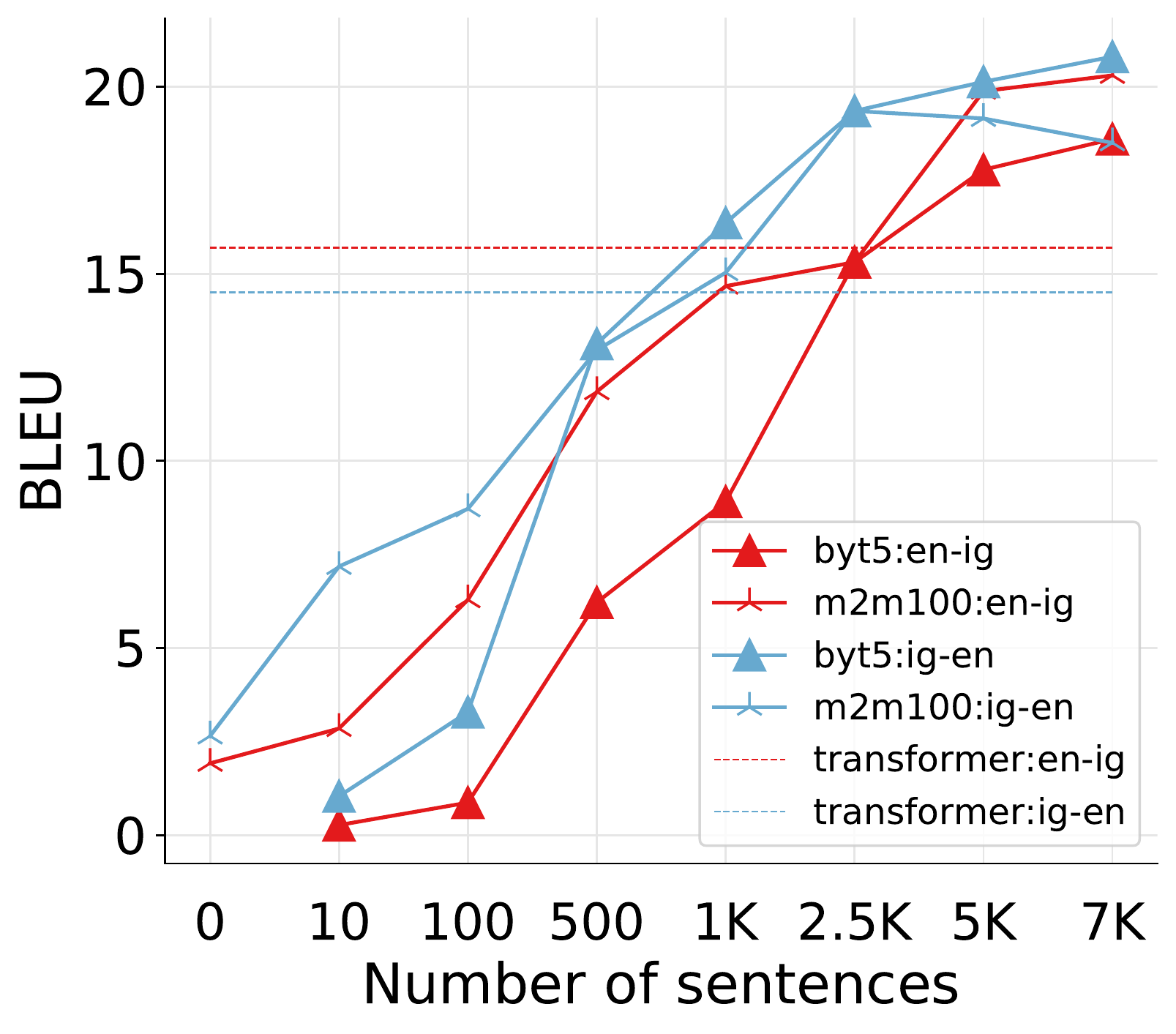}
        \caption{eng-ibo} \label{fig:ig_n}
    \end{subfigure}
    \hspace{-7.2\baselineskip}
    \begin{subfigure}{0.45\textwidth}
        \centering
        \includegraphics[width=0.6\textwidth]{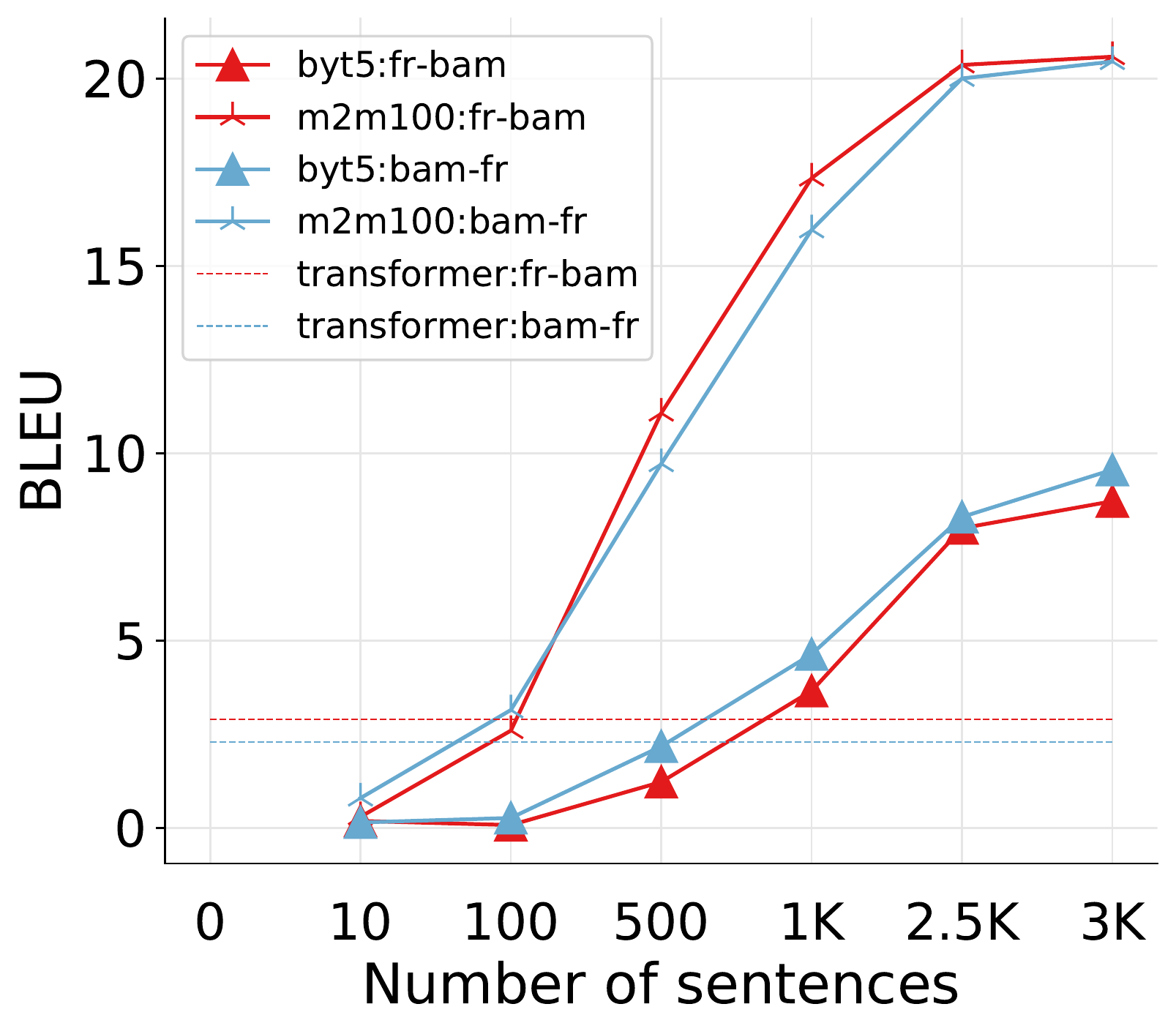}
        \caption{fr-bam} \label{fig:yo_n}
    \end{subfigure}
    \hspace{-3\baselineskip}
    \vspace{-3mm}
    \caption{\textbf{Number of fine-tuning sentences} needed to exceed the performance of a bilingual Transformer model.
    \label{fig:n_sent_analysis}}
    \label{fig:finetuning_size}
\end{figure*}

\begin{table*}[t]
 \footnotesize
 \begin{center}
 \resizebox{\textwidth}{!}{%
  \begin{tabular}{lrrrrrr|rrrrrrrrrr||rr}
    \toprule
     & \multicolumn{6}{c}{\textit{xx-fr}} & \multicolumn{10}{c}{\textit{xx-en}} \\
    \textbf{Model} & \textbf{bam} & \textbf{bbj} & \textbf{ewe} & \textbf{fon} & \textbf{mos} & \textbf{wol} & \textbf{hau} & \textbf{ibo} & \textbf{lug} & \textbf{luo} & \textbf{pcm} & \textbf{swa} & \textbf{tsn} & \textbf{twi} & \textbf{yor} & \textbf{zul}  & \textbf{AVG} & \textbf{MED} \\
    \midrule
    & \multicolumn{11}{c}{\textbf{BLEU}} \\
    \midrule
    
    \multicolumn{2}{l}{Transformer} \\
    \texttt{REL+NEWS} & $4.9$ & $0.6$ & $6.3$ & $2.2$ & $3.7$ & $2.2$ & $11.2$ & $17.4$ & $5.6$ & $3.1$ & $19.5$ & $28.0$ & $23.9$ & $9.8$ & $12.0$ & $27.3$ & $11.1$ & $8.0$ \\
    \texttt{REL$\rightarrow$NEWS} & $4.7$ & $0.8$ & $6.5$ & $2.4$ & $3.1$ & $2.5$ & $11.0$ & $17.4$ & $6.3$ & $1.8$ & $19.0$ & $27.9$ & $24.6$ & $10.1$ & $11.0$ & $28.5$ & $11.1$ & $8.3$ \\
    \texttt{REL+NEWS$\rightarrow$NEWS} & $5.8$ & $1.0$ & $7.1$ & $2.4$ & $\mathbf{4.1}$ & $2.6$ & $13.2$ & $18.2$ & $6.8$ & $3.7$ & $21.4$ & $28.7$ & $24.5$ & $10.4$ & $12.6$ & $30.1$ & $12.0$ & $8.8$ \\
    \midrule
    \multicolumn{2}{l}{M2M-100} \\
    \texttt{REL+NEWS} & $24.0$ & $5.8$ & $10.9$ & $9.7$ & $2.3$ & $10.1$ & $15.3$ & $21.1$ & $21.1$ & $13.3$ & $\mathbf{44.6}$ & $29.4$ & $27.0$ & $12.5$ & $17.4$ & $30.6$ & $18.4$ & $16.4$ \\
    \texttt{REL$\rightarrow$NEWS} & $20.3$ & $5.9$ & $11.4$ & $9.6$ & $2.3$ & $10.5$ & $17.4$ & $\mathbf{21.9}$ & $20.6$ & $13.7$ & $44.3$ & $\mathbf{30.6}$ & $27.7$ & $\mathbf{13.2}$ & $\mathbf{18.0}$ & $36.0$ & $19.0$ & $17.7$ \\
    \texttt{REL+NEWS$\rightarrow$NEWS} & $\mathbf{25.8}$ & $\mathbf{6.3}$ & $\mathbf{11.6}$ & $\mathbf{9.9}$ & $2.6$ & $\mathbf{11.5}$ & $\mathbf{18.2}$ & $21.5$ & $\mathbf{22.4}$ & $\mathbf{14.3}$ & $44.0$ & $30.5$ & $\mathbf{27.8}$ & $\mathbf{13.2}$ & $\mathbf{18.0}$ & $\mathbf{38.1}$ & $\mathbf{19.7}$ & $\mathbf{18.1}$ \\
    
    \bottomrule 
    \toprule 
    & \multicolumn{11}{c}{\textbf{CHRF}} \\
    \midrule
    \multicolumn{2}{l}{Transformer} \\
    \texttt{REL+NEWS} & $24.7$ & $12.6$ & $29.4$ & $16.1$ & $\mathbf{17.6}$ & $19.9$ & $31.7$ & $43.1$ & $26.9$ & $23.0$ & $47.8$ & $53.5$ & $49.8$ & $34.4$ & $33.4$ & $49.6$ & $32.1$ & $30.6$ \\
    \texttt{REL$\rightarrow$NEWS} & $23.0$ & $12.7$ & $29.8$ & $16.6$ & $17.2$ & $18.3$ & $30.6$ & $42.8$ & $28.7$ & $20.0$ & $47.3$ & $53.3$ & $50.8$ & $34.4$ & $32.2$ & $50.4$ & $31.8$ & $30.2$ \\
    \texttt{REL+NEWS$\rightarrow$NEWS} & $26.5$ & $14.7$ & $30.7$ & $17.6$ & $18.8$ & $21.8$ & $33.8$ & $44.0$ & $29.5$ & $24.7$ & $50.8$ & $54.1$ & $50.6$ & $35.1$ & $34.4$ & $51.4$ & $33.7$ & $32.2$ \\
    \midrule
    \multicolumn{2}{l}{M2M-100} \\
    \texttt{REL+NEWS} & $47.1$ & $27.5$ & $36.4$ & $27.9$ & $16.6$ & $34.0$ & $36.8$ & $47.5$ & $47.2$ & $37.3$ & $\mathbf{68.9}$ & $54.7$ & $53.0$ & $38.4$ & $40.2$ & $53.3$ & $41.7$ & $39.3$ \\
    \texttt{REL$\rightarrow$NEWS} & $44.5$ & $27.7$ & $37.0$ & $28.2$ & $16.8$ & $34.4$ & $39.6$ & $\mathbf{48.0}$ & $47.0$ & $38.0$ & $68.7$ & $\mathbf{55.8}$ & $53.6$ & $\mathbf{38.7}$ & $40.7$ & $56.4$ & $42.2$ & $40.2$ \\
    \texttt{REL+NEWS$\rightarrow$NEWS} & $\mathbf{49.0}$ & $\mathbf{28.5}$ & $\mathbf{37.2}$ & $\mathbf{28.9}$ & $17.2$ & $\mathbf{35.3}$ & $\mathbf{40.2}$ & $47.9$ & $\mathbf{48.5}$ & $\mathbf{38.3}$ & $68.6$ & $55.7$ & $\mathbf{54.0}$ & $\mathbf{38.7}$ & $\mathbf{41.0}$ & $\mathbf{57.7}$ & $\mathbf{42.9}$ & $\mathbf{40.6}$ \\
\midrule
    \bottomrule
  \end{tabular}
  }
  \vspace{-1mm}
  \caption{\textbf{Results adapting to Domain Shift, xx-en/fr}. We calculate BLEU and ChrF on the news domain when training on different combinations of \texttt{REL} and \texttt{NEWS}.}
  \label{tab:rel_news_xx_fr_en}
  \end{center}
\end{table*}
To improve over the baseline performance on \texttt{NEWS}, we train bilingual Transformer models (as a baseline) and M2M-100 on a combination of \texttt{REL} and \texttt{NEWS}. We chose M2M-100 because it was the best performing model. \autoref{tab:rel_news_fr_en_xx} gives the BLEU on three settings: \texttt{REL+NEWS}, \texttt{REL$\rightarrow$NEWS}, and \texttt{REL+NEWS$\rightarrow$NEWS}. In general, the improvement depends on the size of REL corpus.
For languages trained on the Bible such as \texttt{bbj}, \texttt{bam}, \texttt{lug}, \texttt{luo}, and \texttt{wol}, the improvement is minimal. 
For M2M-100, the \texttt{REL+NEWS} performance does not improve over \texttt{NEWS} despite the larger quantity of training data. 
This demonstrates that increasing the size in the target domain is the most helpful strategy (see \autoref{fig:finetuning_size}). 
Similarly, combining \texttt{REL+NEWS} is not very helpful for xx-en/fr.
An alternative approach is \texttt{REL$\rightarrow$NEWS}, which allows the model to develop a good understanding of the desired language before adapting to the news domain. We observe an increase on $1.1$ BLEU over \texttt{REL+NEWS} in the en/fr-xx direction. 
However, the best strategy is \texttt{REL+NEWS$\rightarrow$NEWS}, especially for  xx-en/fr where it yields an improvement over \texttt{NEWS} and \texttt{REL+NEWS} by $2.0$ and $1.5$ BLEU, respectively. 







\begin{figure}[t]
    \vspace{-6mm}
    \centering
    \includegraphics[width=0.89\linewidth]{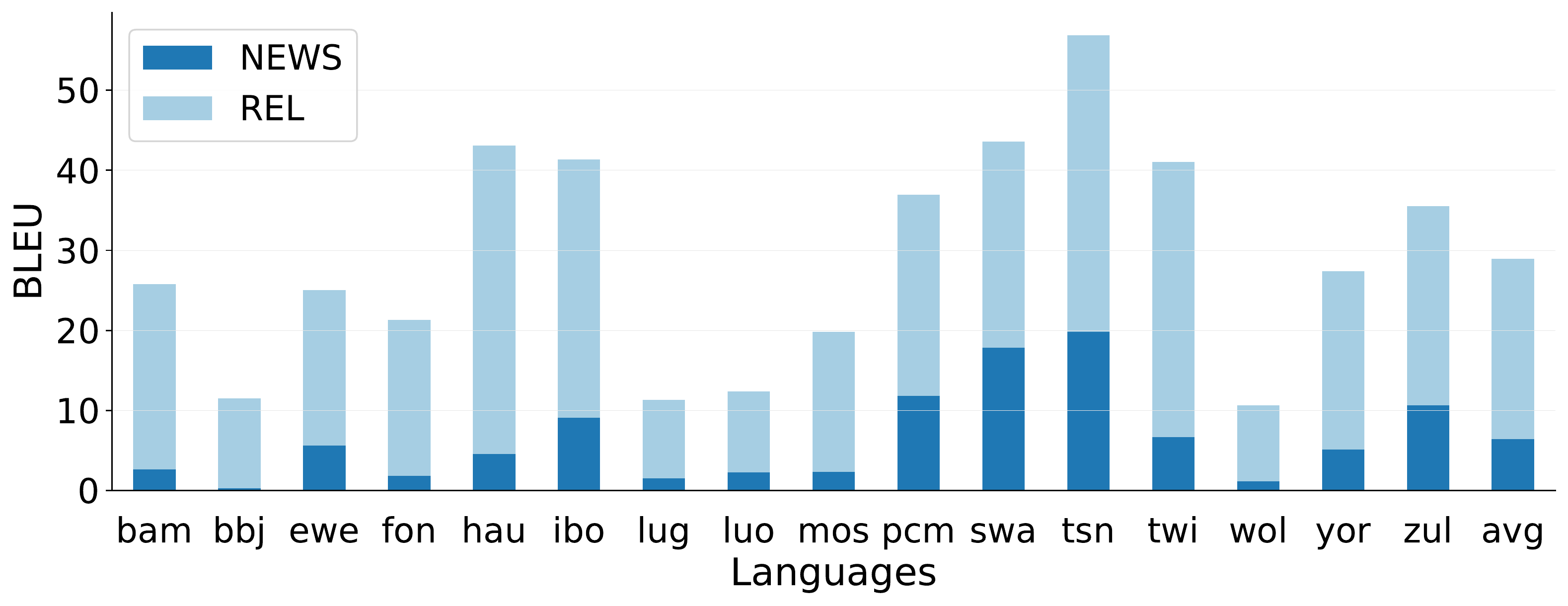}
    \includegraphics[width=0.89\linewidth]{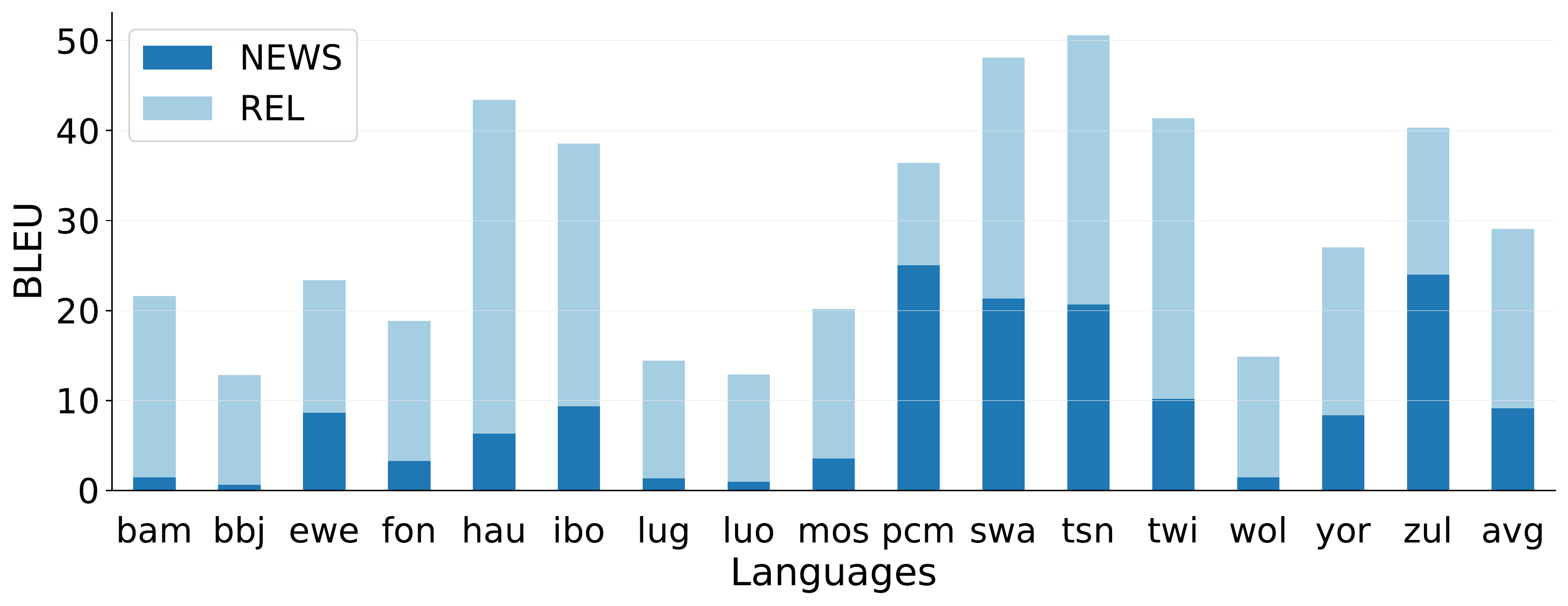}
    \vspace{-2mm}
    \caption{\textbf{Domain shift} of M2M-100 Transformer models trained on en/fr-xx (top) or xx-en/fr (bottom) \texttt{REL} domain and tested on the \texttt{NEWS} vs. \texttt{REL} domains.}
    \label{fig:domain_shift}
    
\end{figure}

 \begin{table}[ht]
 \vspace{-3mm}
 \footnotesize
 \begin{center}
 \scalebox{0.80}{
  \begin{tabular}{lp{75mm}}
   \toprule
    \textit{bam-fr} \\
    \texttt{SRC} & Ani k’a f\textopeno u ye ko c{\textepsilon}manc{\textepsilon} fanga b{\textepsilon} sigi ntuloma saba kan. \\
    \texttt{TGT} &  Et leur dire que la transition se repose sur trois \textcolor{blue}{piliers}. \\
    \texttt{REL} & Et qu'on leur dise que la puissance du milieu est sur trois \textcolor{red}{sauterelles}; \\
    \texttt{R+N$\rightarrow$N} & Et de leur dire que la force de la transition repose sur trois \textcolor{blue}{piliers}. \\
    \midrule
    \textit{lug-en} \\
     \texttt{SRC} & Murasaki Shikibu yawandiika ekitabo ekijjuvu akaasookera ddala mu nsi yonna. \\
    \texttt{TGT} & Murasaki Shikibu wrote \textcolor{blue}{the world's} first full \textcolor{blue}{novel}. \\
    \texttt{REL} & And Murshach Shikib writes a full \textcolor{red}{scroll} of the first \textcolor{red}{in all the earth}. \\
    \texttt{R+N$\rightarrow$N} &  Murasaki Shikibu wrote a complete \textcolor{blue}{book} first \textcolor{blue}{in the world}. \\
    \bottomrule
  \end{tabular}
 }
 \vspace{-3mm}
 \caption{\textbf{Example translations} for M2M-100 fine-tuned on \texttt{REL} or \texttt{REL+NEWS$\rightarrow$NEWS} (\texttt{R+N$\rightarrow$N}). Terms in \textcolor{red}{red} are typical for biblical texts, while the terms in \textcolor{blue}{blue} are more neutral expressions.} 
 \vspace{-7mm}
 
  \label{tab:qual_examples}
  \end{center}
\end{table}

\subsection{Analysis of Domain Shift}


\label{sec:domain_analysis}
\paragraph{Is a small in-domain set essential for fine-tuning?} If we train models \emph{only} on previously available religious data, they are not capable of translating news well due to the strong \textit{domain bias}. This is illustrated in \autoref{fig:domain_shift}: All models perform much worse on \texttt{NEWS} than on the \texttt{REL} domain. 
When the quantity of religious training data is small, the loss in translation performance on the news test set is largest, c.f. \texttt{bbj} (8k of \texttt{REL} data) with a drop of -95.5\% BLEU or \texttt{bam} (-93.5\%, 28k) and \texttt{luo} (-93.5\%, 31k).
This indicates that when the \texttt{REL} training data  is sparse, 
it is insufficient to teach the M2M-100 model a more general understanding required for translating \texttt{NEWS}.
However, when the religious training data is larger, this loss is reduced, c.f. when translating to \texttt{zul} (667k, -67\%), \texttt{swa} (-69.3\%, 872k), and \texttt{tsn} (-71\%, 870k).
While this is the general trend, \texttt{pcm}, whose religious training data is small (23k), has the lowest drop in performance (-59.3\%), which may be due to the strong similarity to its source language.

\paragraph{How many sentences in the target domain are required?} \autoref{fig:finetuning_size} shows how for three selected language pairs with 
a large (\texttt{fr-bam}), medium (\texttt{eng-ibo}) and relatively small (\texttt{eng-swa}) domain gap, the quality of target domain translations improves as we increase the size of the target domain corpus. For all three pairs, fine-tuning M2M-100 or ByT5 on 2.5$k$ sentence pairs of in-domain data (\texttt{NEWS}) is sufficient to outperform the bilingual Transformer baselines that were additionally trained on larger amounts of out-of-domain data (\texttt{REL}). Surprisingly, this procedure not only works for languages included during pre-training (\texttt{swa}), but also for previously unseen languages (\texttt{ibo}, \texttt{bam}). 
M2M-100 tends to adapt to the new data more quickly than ByT5, but in all cases, models continue to learn with additional in-domain data. 
This shows how much more effectively a small number of in-domain translations can be used when they serve for fine-tuning multilingual pre-trained models rather than training bilingual MT models from scratch.

\paragraph{Examples of Domain Bias.}
To illustrate the challenge of overcoming domain bias, we show
 examples translating from \texttt{bam} and \texttt{lug} in \autoref{tab:qual_examples}. The M2M-100 model fine-tuned only on \texttt{REL} succeeds in roughly capturing the meaning of the sources, but using biblical terms, such as ``scroll'' instead of ``novel''. Adding our news corpus to fine-tuning resolves these issues (e.g. ``book'').

\begin{table}[t]
 \footnotesize
 \begin{center}
 \scalebox{0.70}{
  \begin{tabular}{p{13mm}p{8mm}rrrrrrrr}
    \toprule
     & \textbf{Tuned} & & & & & & & & \\
   \textbf{Evaluation} & \textbf{on} & & & & & & & & \\
     \textbf{Domain} & \textbf{NEWS} & \textbf{hau} & \textbf{ibo} & \textbf{lug} & \textbf{luo} & \textbf{swa} & \textbf{wol}  & \textbf{yor}  & \textbf{zul} \\
    \midrule
    \textit{en/fr-xx} \\ 
    FLORES & \xmark &$2.6$ & $2.8$ & $0.8$ & $-$  & $20.9$ & $0.6$ & $1.5$ & $3.3$   \\
    FLORES & \cmark &$4.0$ & $19.9$ & $7.6$ & $13.7$  & $27.1$ & $8.2$ & $13.4$ & $19.2$   \\
    REL & \xmark &$1.2$ & $1.0$ & $0.0$ & $-$  & $11.0$ & $0.0$ & $0.4$ & $1.6$   \\
    REL & \cmark &$3.7$ & $10.3$ & $3.3$ & $5.4$  & $14.6$ & $6.7$ & $10.6$ & $13.0$   \\
    \midrule
    \textit{xx-en/fr} \\ 
    FLORES & \xmark & $8.0$ & $7.2$ & $3.7$ & $-$  & $26.9$ & $3.0$ & $3.8$ & $11.9$   \\

    FLORES & \cmark & $16.3$ & $12.0$ & $7.7$ & $11.8$  & $25.8$ & $7.5$ & $9.3$ & $19.2$   \\
   REL & \xmark &$6.4$ & $3.7$ & $0.5$ & $-$  & $15.4$ & $0.4$ & $0.9$ & $8.5$   \\
    REL & \cmark &$3.8$ & $6.0$ & $1.7$ & $2.5$  & $13.9$ & $1.7$ & $5.7$ & $12.5$   \\
    \bottomrule
  \end{tabular}
 }
 \caption{\textbf{spBLEU on Wikipedia domain} (FLORES) and REL for M2M-100 before (\xmark) and after (\cmark) fine-tuning on \texttt{NEWS}. }
 \vspace{-3mm}
  \label{tab:dshift_flores}
  \end{center}
\end{table}

\paragraph{How general is our news corpus?}
\autoref{tab:dshift_flores} shows the zero-shot evaluation of M2M-100 fine-tuned on our small \texttt{NEWS} corpora on other domains: religious (\texttt{REL}) and Wikipedia (FLORES). We evaluated the Wikipedia domain on the FLORES \textit{devtest} and the \texttt{REL} domain on either JW300 or Bible (\texttt{lug}, \texttt{luo}, \texttt{wol}). As a baseline, we evaluated the zero-shot performance of M2M-100 (not fine-tuned, \xmark) on FLORES\footnote{except for Luo which is not supported} using spBLEU (i.e. sentencepiece BLEU~\cite{Goyal2021TheFE}). We noticed very poor performance except for Swahili --- as discussed in \S\ref{sec:adapt_focus_lang}.
After fine-tuning on our new data (\cmark), transfer is largely improved across the bench (up to +17 BLEU for \texttt{en-ibo}). The same trend holds for the religious domain. This shows that even though our data comes from the news domain, it helped the model generalize to other domains.
Hence, expanding African news corpora and
developing better MT models for news pays off even for other domains of interest.



\section{Conclusion}

We have created MAFAND-MT, a corpus of 16 African languages to study  translation systems for low-resource languages in the news domain. 
We investigate how to most effectively adapt large-scale pre-trained models to incorporate new languages and new domains.
Our findings suggest that as little as 2k sentences are sufficient for fine-tuning, with an improved performance, paving the way for others to create new translation systems without relying on large collections of web-sourced text. This has strong implications for languages that are spoken by millions but lack presence on the web.   

In the future, we hope to expand our coverage to additional under-resourced languages, and to develop even more effective fine-tuning objectives. Currently, we are extending our corpus to Amharic, Chichewa, Kinyarwanda, Shona, and isiXhosa, including an expansion of the Hausa corpus,  they will be released under MAFAND-MT dataset name\footnote{We provide details on the evaluation datasets in \autoref{sec:new_eval_dataset}}.

\section{Acknowledgment}
This work was carried out with support from Lacuna Fund, an initiative co-founded by The Rockefeller Foundation, Google.org, and Canada’s International Development Research Centre. David Adelani acknowledges the EU-funded Horizon 2020 projects: COMPRISE (\texttt{http://www.compriseh2020.eu/}) under grant agreement No. 3081705 and ROXANNE under grant number 833635. We thank Chester Chester Palen-Michel and Constantine Lignos for providing the VOA corpus for this research, and Google for providing GCP credits to run some of the experiments. Finally, we thank Davor Orlič and Knowledge4All for their administrative support throughout the project. 

\bibliography{anthology,custom}
\bibliographystyle{acl_natbib}

\appendix

\section{Language Characteristics}
\autoref{tab:lang_char} provides the details about the language characteristics. 
\label{sec:appendix_lang_char}
 \begin{table*}[t]
 \footnotesize
 \begin{center}
 \resizebox{\textwidth}{!}{%
  \begin{tabular}{lrlp{50mm}llll}
    \toprule
     & \textbf{No. of} &\textbf{Latin Letters} & \textbf{Letters}  &  &  &  \textbf{sentence}\\
    \textbf{Language} & \textbf{Letters} & \textbf{Omitted} & \textbf{added}  & \textbf{Tonality} & \textbf{diacritics} & \textbf{morphology} & \textbf{structure}\\
    \midrule
    Bambara (bam) & 27 & q,v,x & \textepsilon, \textopeno, \textltailn, \textipa{\ng} &  yes, 2 tones & yes & isolating & SVO \& SOV \\
    \ghomala (bbj) & 40 & q, w, x, y & bv, dz, \textschwa, a\textschwa, \textepsilon, gh, ny, nt, \textipa{\ng}, \textipa{\ng}k, \textopeno, pf, mpf, sh, ts, \textbaru, zh, '  &  yes, 5 tones & yes & agglutinative & SVO \\
    \ewe (ewe) & 35  & c, j, q & \textrtaild, dz, \textepsilon, \textflorin, gb, \textgamma, kp, ny, \textipa{\ng}, \textopeno, ts, \textscriptv   & yes, 3 tones  & yes & isolating & SVO \\
    Fon (fon) & 33 & q & \textrtaild, \textepsilon,gb, hw, kp, ny, \textopeno, xw & yes, 3 tones & yes  & isolating & SVO \\
    Hausa (hau) & 44 & p,q,v,x & \texthtb, \texthtd, \texthtk, \begin{tfour}\m{y}\end{tfour}, kw, {\texthtk}w, gw, ky, {\texthtk}y, gy, sh, ts  & yes, 2 tones  & no & agglutinative & SVO \\
    Igbo (ibo) & 34 & c, q, x & ch, gb, gh, gw, kp, kw, nw, ny, {\d o}, \.{o}, sh, {\d u}& yes, 2 tones  & yes & agglutinative & SVO \\
    Luganda (lug) & 25 & h, q, x & \textipa{\ng}, ny & yes, 3 tones & no & agglutinative & SVO \\
    Luo (luo) & 31 & c, q, x, v, z &   ch, dh, mb, nd, ng’, ng, ny, nj, th, sh & yes, 4 tones & no & agglutinative & SVO \\
    Mossi (mos) & 26 & c, j, q, x  & ', \textepsilon, \textiota, \textscriptv & yes, 2 tones  & yes  & isolating & SVO \\
    Naija (pcm) & 26 & -- & -- & no  & no & mostly analytic & SVO \\
    Swahili (swa) & 33 & x, q & ch, dh, gh, kh, ng', ny, sh, th, ts & no & yes   & agglutinative & SVO \\
    Setswana (tsn) & 36 & c, q, v, x, z & \^{e}, kg, kh, ng, ny, \^o, ph, \v{s}, th, tl, tlh, ts, tsh, t\v{s}, t\v{s}h & yes, 2 tones & no & agglutinative & SVO \\
    Akan/Twi (twi) & 22 & c,j,q,v,x,z & \textepsilon, \textopeno & yes, 5 tones  & no & isolating & SVO \\
    Wolof (wol) &29& h,v,z & \textipa{\ng}, \`a, \'e, \"{e}, \'o, \~{n}  & no & yes  & agglutinative & SVO \\
    \yoruba (yor) & 25 & c, q, v, x, z & {\d e}, gb, {\d s} , {\d o} & yes, 3 tones & yes  & isolating & SVO \\
    \zulu (zul) & 55 & -- & nx, ts, nq, ph, hh, ny, gq, hl, bh, nj, ch, ngc, ngq, th, ngx, kl, ntsh, sh, kh, tsh, ng, nk, gx, xh, gc, mb, dl, nc, qh &  yes, 3 tones & no & agglutinative & SVO \\
    \bottomrule
  \end{tabular}
  }
  \vspace{-3mm}
  \caption{Linguistic Characteristics of the Languages}
  \label{tab:lang_char}
  \end{center}
\end{table*}

\section{Available Parallel Corpora}
\label{sec:avail_paral_corpus}
We found Five African languages with publicly available parallel texts in the news domain: Hausa, Igbo, Swahili, \yoruba, and \zulu. \autoref{tab:data_stat} provides news source, the \texttt{TRAIN}, \texttt{DEV} and \texttt{TEST} splits. 
\paragraph{Hausa} The Hausa Khamenei\footnote{\url{https://www.statmt.org/wmt21/translation-task.html}} corpus contains 5,898 sentences, we split them into \texttt{TRAIN} (3,098), \texttt{DEV} (1,300), and \texttt{TEST} split (1,500). 
\paragraph{Igbo} The Igbo corpus~\cite{Ezeani2020IgboEnglishMT} has 9,998 sentences, we extract 6,998 sentences for \texttt{TRAIN}, and the remaining for \texttt{DEV} and \texttt{TEST} splits. 
\paragraph{Swahili} The Global Voices\footnote{\url{https://sw.globalvoices.org/}} corpus contains 30,782 sentences, which we use for
the \texttt{TRAIN} split. We additionally crawled newer (2019--2021) publications of Swahili articles from the Global Voices website, this gives a total of 3,626 sentences, they were aligned and manually verified by Swahili speakers. They are 
split into the \texttt{DEV} and \texttt{TEST} splits. 

\paragraph{\yoruba} The MENYO-20k~\cite{adelani-etal-2021-effect} corpus contains sentences from different domains (TED talks, books, 
software localization, proverbs, and news), from which we select the news domain sentences for the \texttt{TRAIN}, \texttt{DEV} and \texttt{TEST} splits.  

\paragraph{\zulu} The Umsuka corpus~\cite{rooweither_mabuya_2021_5035171} contains 9,703 training sentences and 1,984 evaluation sentences. 4,739 training sentences were translated from English-\zulu, and the remaining from \zulu-English. We only keep the training sentences translated into \zulu, and split them into 3,500 for \texttt{TRAIN} and 1,239 sentences for \texttt{DEV}. 
From the existing evaluation set we select only the 998 English-\zulu translations for \texttt{TEST}. Umsuka provides two translations for each English sentence, but we use only the first. 

\section{Monolingual Corpus PLMs adaptation}
\label{sec:monolingual_corpus}

\begin{table*}[t]
\footnotesize
 \begin{center}
 \resizebox{\textwidth}{!}{%
\begin{tabular}{llrr}

\toprule
\textbf{ Language} & \textbf{Source} & \textbf{Size (MB)} & \textbf{No. of sentences}  \\
\midrule
Afrikaans (afr) & mC4 (subset)~\cite{xue-etal-2021-MT5} & 752.2MB & 3,697,430\\
Amharic (amh) & mC4 (subset), and VOA & 1,300MB & 2,913,801\\
Arabic (ara) & mC4 (subset) & 1,300MB & 3,939,375\\
English (eng) & mC4 (subset), and VOA & 2,200MB & 8,626,571 \\
French (fra) & mC4 (subset), and VOA & 960MB & 4,731,196 \\
Hausa (hau) & mC4 (all), and VOA & 594.1MB & 3,290,382\\
Igbo (ibo) & mC4 (all), and AfriBERTa Corpus~\cite{ogueji-etal-2021-small} & 287.5MB & 1,534,825\\
Malagasy (mg) & mC4 (all) & 639.6MB & 3,304,459 \\
Chichewa (nya) & mC4 (all), Chichewa News Corpus~\cite{Siminyu2021AI4DA} & 373.8MB & 2,203,040 \\
Oromo (orm) & AfriBERTa Corpus, and VOA & 67.3MB & 490,399\\
Naija (pcm) & AfriBERTa Corpus, and VOA & 54.8MB & 166,842\\
Rwanda-Rundi (kir/kin) & AfriBERTa Corpus, KINNEWS \& KIRNEWS~\cite{niyongabo-etal-2020-kinnews}, and VOA & 84MB & 303,838\\
Shona (sna) & mC4 (all), and VOA & 545.2MB & 2,693,028\\
Somali (som) & mC4 (all), and VOA & 1,000MB & 3,480,960 \\
Sesotho (sot) & mC4 (all) & 234MB & 1,107,565 \\
Swahili (swa) & mC4 (all) & 823.5MB & 4,220,346 \\
isiXhosa (xho) & mC4 (all), and Isolezwe Newspaper  & 178.4MB &  832,954 \\
\yoruba (yor) & mC4 (all), Alaroye News, Asejere News, Awikonko News, BBC, and VON~\cite{adelani-etal-2021-masakhaner}  & 179.3MB & 897,299 \\
\zulu (zul) & mC4 (all), and Isolezwe Newspaper & 700.7MB & 3,252,035  \\
\bottomrule 

\end{tabular}
}
\footnotesize
  \caption{Monolingual Corpora (after pre-processing -- we followed AfriBERTa~\cite{ogueji-etal-2021-small} approach) , their sources and size (MB), and number of sentences. }
  \label{tab:monolingual_corpus}
\end{center}
\end{table*}

\autoref{tab:monolingual_corpus} provides the details about the Monolingual corpus used to adapt the pre-trained language models (PLMs), their size and source of corpora. The African languages pre-trained are: Afrikaans, Amharic, Hausa, Igbo, Malagasy, Chichewa, Oromo, Naija, Kinyarwanda, Kirundi, Shona, Somali, Sesotho, Swahili, isiXhosa, \yoruba, and \zulu.

\label{sec:appendix_lang_char}
 \begin{table*}[t]
 \footnotesize
 \begin{center}
 \resizebox{\textwidth}{!}{%
  \begin{tabular}{llp{100mm}}
    \toprule
     \textbf{Model Name} &\textbf{HuggingFace Model name} & \textbf{Remark} \\
    \midrule
    AfriMT5 & \url{masakhane/afri-mt5-base} & mT5-base adaptation to 17 African languages, English, French and Arabic. \\
    AfriByT5 & \url{masakhane/afri-byt5-base} & ByT5-base adaptation to 17 African languages, English, French and Arabic. \\
    AfriMBART & \url{masakhane/afri-mbart50} & mBART50 adaptation to 17 African languages, English, French and Arabic. \\
     \midrule
     
    \texttt{NEWS} (MT5) & \url{masakhane/mt5_{src}_{tgt}_news} & MT5 fine-tuned on \{src\}-\{tgt\} direction using parallel NEWS corpus. \\
    \texttt{NEWS} (AfriMT5) & \url{masakhane/afrimt5_{src}_{tgt}_news} & AfriMT5 fine-tuned on \{src\}-\{tgt\} direction using parallel NEWS corpus. \\
    
    \texttt{NEWS} (ByT5) & \url{masakhane/byt5_{src}_{tgt}_news} & ByT5 fine-tuned on \{src\}-\{tgt\} direction using parallel NEWS corpus. \\
    \texttt{NEWS} (AfriByT5) & \url{masakhane/afribyt5_{src}_{tgt}_news} & AfriByT5 fine-tuned on \{src\}-\{tgt\} direction using parallel NEWS corpus. \\
    
    \texttt{NEWS} (mBART50) & \url{masakhane/mbart50_{src}_{tgt}_news} & mBART50 fine-tuned on \{src\}-\{tgt\} direction using parallel NEWS corpus. \\
    \texttt{NEWS} (AfriByT5) & \url{masakhane/afrimbart_{src}_{tgt}_news} & AfriMBART fine-tuned on \{src\}-\{tgt\} direction using parallel NEWS corpus. \\
    \texttt{NEWS} (M2M-100) & \url{masakhane/m2m100_418M_{src}_{tgt}_news} & M2M-100 fine-tuned on \{src\}-\{tgt\} direction using parallel NEWS corpus. \\
    \midrule
    \texttt{NEWS} (M2M-100-EN) & \url{masakhane/m2m100_418M-EN-NEWS} & M2M-100 fine-tuned on NEWS data that are English-centric i.e \texttt{en}--\{\texttt{hau}, \texttt{ibo}, \texttt{lug}, \texttt{luo}, \texttt{pcm}, \texttt{swa}, \texttt{tsn}, \texttt{twi}, \texttt{yor}, \texttt{zul}\} \\
    \texttt{NEWS} (M2M-100-FR) & \url{masakhane/m2m100_418M-FR-NEWS} & M2M-100 fine-tuned on NEWS data that are French-centric i.e \texttt{fr}--\{\texttt{bam}, \texttt{bbj}, \texttt{ewe},  \texttt{fon}, \texttt{mos}, \texttt{wol}\}.  \\
    \midrule
      
    \texttt{REL} & \url{masakhane/m2m100_418M_{src}_{tgt}_rel} & M2M-100 fine-tuned on \{src\}-\{tgt\} direction using parallel REL corpus.  \\
     
    \texttt{REL+NEWS} & \url{masakhane/m2m100_418M_{src}_{tgt}_rel_news} & M2M-100 fine-tuned on \{src\}-\{tgt\} direction using parallel REL+NEWS corpus.  \\
    
    \texttt{REL$\rightarrow$NEWS} & \url{masakhane/m2m100_418M_{src}_{tgt}_rel_ft} & M2M-100 fine-tuned on \{src\}-\{tgt\} direction using parallel REL corpus and additional fine-tuning on NEWS  \\
    \texttt{REL+NEWS$\rightarrow$NEWS} & \url{masakhane/m2m100_418M_{src}_{tgt}_rel_news_ft} & M2M-100 fine-tuned on \{src\}-\{tgt\} direction using parallel REL+NEWS and additional fine-tuning on NEWS  \\
    \bottomrule
  \end{tabular}
  }
  \vspace{-3mm}
  \caption{Model names on HuggingFace Model Hub. For bilingual models, supply the correct \textbf{src} or \textbf{tgt} language. English/French make use of a 2-letter language code i.e en or fr, while all the African languages make us of 3-letter language codes e.g yor.}
  \label{tab:model_names}
  \end{center}
\end{table*}

\section{Model Hyper-parameters and 
Reproducibility of Results}
\label{sec:hyperparameter}
For the pre-trained models, we fine-tune the models using HuggingFace transformer tool ~\cite{wolf-etal-2020-transformers} with the default learning rate ($5e-5$), batch size of $10$, maximum source length \& maximum target length of $200$, beam size of $10$, and number of epochs is $3$ except for models trained on only \texttt{NEWS} which we set to $10$. We make All the experiments were performed on
a single GPU (Nvidia V100). 

For fine-tuning pre-trained models, especially for mBART50 that only supports two African languages, the target language is required to
be specified during decoding from among those
that the model has seen during pre-training, we follow past works~\cite{madaan-etal-2020-transfer,cahyawijaya-etal-2021-indonlg,Lee2022PreTrainedMS} in selecting another closely-related language that is represented in the pre-trained model. For convenience, we make use of Swahili (sw) as the target language when an African language is not represented since Swahili is represented in all the pre-trained models. The only exception is Nigerian-Pidgin, where we make use of French (fr) since it is closely related to English. When a language is represented in the pre-trained model like M2M-100 has seen \yoruba (yo), we make use of the correct language code. 

To train AfriMT5 and ByT5, we start with MT5 and ByT5.
We pre-train with the learning rate 
$1e-4$, $10,000$ warm up steps and a batch size of $2048$ for one epoch. For mBART50, we pre-train with learning rate of $5e-5$ for $50,000$ steps using Fairseq~\cite{ott-etal-2019-fairseq} without modifying the mBART50 vocabulary. \autoref{tab:model_names} has the names of all the models that are publicly available on HuggingFace Model Hub~\footnote{\url{https://huggingface.co/masakhane}}. In total, we have 357 models from 22 x 16 bilingual models, two English/French-centric models, and three adapted models to African languages (i.e AfriMT5, AfriByT5, and AfriMBART). 

\section{BLEU vs spBLEU}
\label{sec:bleu_spbleu}
\autoref{tab:dshift_flores_bleu} and \autoref{tab:dshift_flores_spbleu} compares BLEU and spBLEU metric for the domain transfer experiments. We observe that spBLEU gives higher scores than BLEU especially in the direction of \textit{en/fr-xx}, which shows that it may be better for evaluating African languages. Although, further analysis and human evaluation are still needed to show that spBLEU is generally better. On the other hand, in the \textit{xx-en/fr}, there is no much difference in the scores between BLEU and spBLEU. 



 \begin{table}[ht]
 \footnotesize
 \begin{center}
 \scalebox{0.70}{
  \begin{tabular}{llrrrrrrrr}
    \toprule
     & \textbf{Tuned} & & & & & & & & \\
   \textbf{Evaluation} & \textbf{on} & & & & & & & & \\
     \textbf{Domain} & \textbf{NEWS} & \textbf{hau} & \textbf{ibo} & \textbf{lug} & \textbf{luo} & \textbf{swa} & \textbf{wol}  & \textbf{yor}  & \textbf{zul} \\
    \midrule
    \textit{en/fr-xx} \\ 
    FLORES & \xmark & $2.4$ & $2.0$ & $0.9$ & $-$  & $19.6$ & $0.4$ & $1.0$ & $1.9$   \\
    FLORES & \cmark & $2.9$ & $12.3$ & $4.9$ & $8.8$  & $22.5$ & $4.2$ & $5.1$ & $8.4$   \\
    REL & \xmark & $2.5$ & $1.8$ & $0.0$ & $-$  & $14.6$ & $0.0$ & $1.4$ & $2.1$   \\
    REL & \cmark & $6.7$ & $9.4$ & $1.1$ & $2.4$  & $17.4$ & $2.7$ & $8.2$ & $8.3$   \\
    NEWS & \xmark & $0.4$ & $2.4$ & $1.8$ & $-$  & $20.1$ & $1.3$ & $2.1$ & $5.6$   \\
    NEWS & \cmark & $14.4$ & $20.3$ & $13.0$ & $10.8$  & $27.0$ & $11.1$ & $12.8$ & $16.5$   \\
    \midrule
    \textit{xx-en/fr} \\ 
    FLORES & \xmark & $6.6$ & $6.0$ & $2.6$ & $-$  & $26.2$ & $2.1$ & $2.7$ & $10.5$   \\
    FLORES & \cmark & $5.4$ & $11.8$ & $6.9$ & $10.3$  & $25.4$ & $6.6$ & $7.9$ & $18.1$   \\
    REL & \xmark & $9.7$ & $5.9$ & $0.5$ & $-$  & $22.3$ & $0.3$ & $1.8$ & $7.8$   \\
   REL & \cmark & $7.7$ & $10.7$ & $1.8$ & $2.6$  & $20.5$ & $1.7$ & $8.8$ & $12.9$   \\
   NEWS & \xmark & $2.2$ & $6.4$ & $4.8$ & $-$  & $25.2$ & $0.8$ & $3.0$ & $13.8$   \\
    NEWS & \cmark & $17.2$ & $18.5$ & $19.4$ & $12.8$  & $29.9$ & $9.5$ & $16.0$ & $36.6$   \\
    \bottomrule
  \end{tabular}
 }
  \vspace{-3mm}
 \caption{\textbf{BLEU on Wikipedia domain} (FLORES), REL, and NEWS for M2M-100 before (\xmark) and after (\cmark) fine-tuning on \texttt{NEWS}. }
  \label{tab:dshift_flores_bleu}
  \end{center}
\end{table}

\begin{table}[t]
 \footnotesize
 \begin{center}
 \scalebox{0.70}{
  \begin{tabular}{llrrrrrrrr}
    \toprule
     & \textbf{Tuned} & & & & & & & & \\
   \textbf{Evaluation} & \textbf{on} & & & & & & & & \\
     \textbf{Domain} & \textbf{NEWS} & \textbf{hau} & \textbf{ibo} & \textbf{lug} & \textbf{luo} & \textbf{swa} & \textbf{wol}  & \textbf{yor}  & \textbf{zul} \\
    \midrule
    \textit{en/fr-xx} \\ 
    FLORES & \xmark &$2.6$ & $2.8$ & $0.8$ & $-$  & $20.9$ & $0.6$ & $1.5$ & $3.3$   \\
    FLORES & \cmark &$4.0$ & $19.9$ & $7.6$ & $13.7$  & $27.1$ & $8.2$ & $13.4$ & $19.2$   \\
    REL & \xmark &$1.2$ & $1.0$ & $0.0$ & $-$  & $11.0$ & $0.0$ & $0.4$ & $1.6$   \\
    REL & \cmark &$3.7$ & $10.3$ & $3.3$ & $5.4$  & $14.6$ & $6.7$ & $10.6$ & $13.0$   \\
    NEWS & \xmark & $0.6$ & $4.1$ & $2.3$ & $-$  & $21.4$ & $1.2$ & $2.4$ & $5.6$   \\
    NEWS & \cmark & $20.2$ & $31.6$ & $22.6$ & $16.4$  & $31.4$ & $19.9$ & $25.5$ & $27.6$   \\
    \midrule
    \textit{xx-en/fr} \\ 
    FLORES & \xmark & $8.0$ & $7.2$ & $3.7$ & $-$  & $26.9$ & $3.0$ & $3.8$ & $11.9$   \\

    FLORES & \cmark & $16.3$ & $12.0$ & $7.7$ & $11.8$  & $25.8$ & $7.5$ & $9.3$ & $19.2$   \\
   REL & \xmark &$6.4$ & $3.7$ & $0.5$ & $-$  & $15.4$ & $0.4$ & $0.9$ & $8.5$   \\
    REL & \cmark &$3.8$ & $6.0$ & $1.7$ & $2.5$  & $13.9$ & $1.7$ & $5.7$ & $12.5$   \\
    NEWS & \xmark & $2.6$ & $9.1$ & $7.2$ & $-$  & $27.8$ & $1.0$ & $3.9$ & $15.7$   \\
    NEWS & \cmark & $17.6$ & $22.8$ & $24.4$ & $15.8$  & $32.0$ & $12.3$ & $17.5$ & $39.0$   \\
    \bottomrule
  \end{tabular}
 }
  \vspace{-3mm}
 \caption{\textbf{spBLEU on Wikipedia domain} (FLORES), REL, and NEWS for M2M-100 before (\xmark) and after (\cmark) fine-tuning on \texttt{NEWS}. }
 \vspace{-4mm}
  \label{tab:dshift_flores_spbleu}
  \end{center}
\end{table}

\section{Qualitative Analysis}\label{sec:appendix_example}
The following examples from the Fon-to-French translations of the test set illustrate the advantage of multilingual modeling and its limitations:
\begin{itemize}
    \item \textbf{Source} (\texttt{fon}):
    Louis Guy Alimanyi{\textrtaild}okpo kp\'od\' Issa Etchlekoun kp\'o \textopeno, s\'in az\v{a}n m{\textopeno}kp\'an {\textrtaild}ye \textopeno, ye \textrtaild\`o w\v{u}v{\textepsilon} s\`e w{\textepsilon} tawun  {\textrtaild}\`o agbaza m\textepsilon, c\'o ye k\'a tuun f\'i \'e az{\textopeno}n n{\textepsilon} l{\textepsilon}{\textepsilon} gosin \'e \textopeno $\,$\v{a}.
    
    \item \textbf{Reference} (\texttt{fr}): Les faits Louis Guy Alimagnidokpo et Issa Etchlekoun se plaignent depuis quelques jours de multiples douleurs, ignorant l’origine réelle de leurs maux.
    \item \textbf{Bilingual Transformer} (\texttt{REL+NEWS},  \texttt{fon}$\rightarrow$\texttt{fr}): on ne peut pas avoir une trentaine d’années ni un jeune homme ni un jeune homme d’âge pour un jeune homme qui soit 12 ans.
    \item \textbf{M2M-100} (\texttt{REL+NEWS$\rightarrow$NEWS}, \texttt{fon}$\rightarrow$\texttt{fr}): Louis Guy Alimanyion et Issa Etchlekoun ont depuis plusieurs jours souffert d’une maladie grave malgré les conséquences de cette maladie qu’ils ne connaissent pas.
    \item \textbf{M2M-100} (\texttt{REL+NEWS$\rightarrow$NEWS}, \texttt{fr}$\rightarrow$\texttt{fon}): S\'in az\v{a}n y{\textopeno}yw{\textepsilon}yw{\textepsilon} {\textrtaild}\'e 
    {\textrtaild}ye {\textrtaild}okpóo w\'{\textepsilon} n\v{u} è kàn Louis Guy Alimagnidokpo kpódó Issa 
    Etchl{\textepsilon}k\'{\textepsilon}n kpán {\textrtaild}è \'\textopeno $\,${\textrtaild}\`o x\'o {\textrtaild}\textopeno 
    $\,$w\'{\textepsilon} {\textrtaild}\'\textopeno 
    $\,$w\v{u}v\'{\textepsilon} gege w\'{\textepsilon}, ye ká 
    tuun nǔ è wú wǔv\'{\textepsilon} yet\textopeno n {\textrtaild}è \'\textopeno $\,$\v{a}.
\end{itemize}
The translation of the bilingual Transformer model is very poor and far from the Fon source, highlighting how poorly the model generalized from the few thousand training sentences. The M2M-100 model gives a more meaningful and adequate translation. M2M-100 makes a surprising but beautiful move, switching \textit{se plaignent depuis quelques jours de multiples douleurs (s\'in az\v{a}n m{\textopeno}kp\'an {\textrtaild}ye \textopeno, ye \textrtaild\`o w\v{u}v{\textepsilon} s\`e w{\textepsilon} tawun  {\textrtaild}\`o agbaza m\textepsilon)} to \textit{ont depuis plusieurs jours souffert d'une maladie grave}. The BLEU score here might be low but the meaning is conserved and even more detailed than the French reference. In fact, in this source context, \textit{wǔv\begin{tfour}\m{e}\end{tfour}} means \textit{souffrir, souffrance (suffer, suffering)}: the French reference made use of \textit{se plaignent (complaining)} which makes less sense than \textit{souffert} used in the M2M-100 prediction. M2M-100 also learned the style of the sentence: \textit{c\'o ye k\'a tuun f\'i \'e az{\textopeno}n n{\textepsilon} l{\textepsilon}{\textepsilon} gosin (but they do know the origin of their sufferings) \'e \textopeno $\,$\v{a} (\textbf{NOT})} - this last part is crucial for the meaning of the entire sentence. Given the structural and morphological differences between Fon and French, we expected it to be more complicated to predict.
However, this translation is structurally wrong even though any French native speaker would understand the conveyed message quickly and easily. In the M2M-100 translation, the word \textit{malgré} is at the wrong place, corrupting syntax and logic of the second clause.
A perfect translation (in the idea to be expressed) would be: "Louis Guy Alimanyion et Issa Etchlekoun ont depuis plusieurs jours souffert d’une maladie grave \sout{malgré} (dont) \textit{ils ne connaissent pas} les \sout{conséquences} (causes/raisons) \sout{de cette maladie qu’ils ne connaissent pas}."

In the opposite translation direction, \texttt{fr}$\rightarrow$\texttt{fon}, M2M-100 (\texttt{REL+NEWS$\rightarrow$NEWS}) still preserved some sense of logical reasoning and predicted the last part right \textit{ye k\'a tuun n\v{u} \`e w\'u w\v{u}v\'{\textepsilon} 
yet\textopeno n (they do know why they are suffering) {\textrtaild}è \'\textopeno $\,$\v{a} (\textbf{NOT})}. However, the model had some limitations: the names which are part of the translation are not spelled correctly. Some expressions are incomplete: For instance \textit{sín az\v{a}n + number} means \textit{since xxx days} but \textit{y{\textepsilon}yw{\textepsilon}} is not a number, and do not have any meaning in this context.

\section{Limitations and Risks}
Despite the promising results, our work has the following limitations:
\begin{enumerate}
    \item \textbf{Translation quality}: Even the best model scores low BLEU on some of the reported languages (\texttt{bbj}, \texttt{mos}, \texttt{zul}), in particular when translating into them. 
    \item \textbf{Evaluation}: Our evaluation is focused on BLEU. We report ChrF results as well, but without a deeper human evaluation, we cannot make claims about the absolute quality of the translations. Manual inspections of translations like the example discussed in Section~\ref{sec:appendix_example} gave us the impression that translations are surprisingly fluent and make good use of language-specific expressions when translating into English or French, but that errors in grammar and logic can be easily overlooked. Automatic reference-based metrics like BLEU and ChrF might not be able to capture the semantic relatedness to the reference sufficiently, as well potentially being tricked by word matches in incoherent phrases. 
    \item \textbf{Language bias}: We have shown that even when not included in pre-training, and without large out-of-domain data, significant gains in translation quality can be achieved. However, language-specific biases, in terms of resourcedness, morphology, standardization, inclusion in pre-trained models and available corpora, or relatedness to other languages, still affect the relative quality of translations, and require more efforts to be overcome.
    \item \textbf{Domain limitations}: While we showed a rapid adaptation to the news domain and the auxiliary benefit of the religious domain, our study also revealed how automatically estimated translation quality drops when the test domain is narrow. Therefore, future work should aim to expand the study to multiple test domains and develop systematic methods for distilling knowledge from multiple narrow domains.
    \item \textbf{Language coverage}: Africa has thousands of other languages that are not covered in our study but deserve the same attention. We hope that our work is encouraging enough to inspire native speakers of those languages not covered here to collect translations, run our code, and report their findings to the NLP research community, so that we can make joint progress in developing language technology for more people.
\end{enumerate}

We believe that our translation models carry similar risks of causing harm by inaccurate and biased translations as the underlying large pre-trained models.  M2M-100 is trained on large collections of texts crawled from the web, and the quality for most of the languages studied here is questionable~\citep{Kreutzer2021QualityAA}. Our fine-tuning successes show that some obvious biases can be overcome when the quality of the fine-tuning set is controlled (see the examples in Section~\ref{sec:domain_analysis}), but we cannot guarantee that biases prevailing in the pre-training corpus or more subtle biases will not occur with other inputs. Together with a careful human evaluation, this should be the main concern for future work on the produced models.
The methodology of rapid fine-tuning might also be misused to tune the models towards harmful content or purposes that harm the speakers of the languages presented here.

\section{New Evaluation Datasets}
\label{sec:new_eval_dataset}

\begin{table*}[ht]
 \footnotesize
 \begin{center}
 \resizebox{\textwidth}{!}{%
  \begin{tabular}{lllr|r|p{52mm}r|lr}
    \toprule
     \textbf{Target}& &\textbf{African} & \textbf{No. of}  & \textbf{Source} & \multicolumn{2}{c|}{\textbf{NEWS}}  & \multicolumn{2}{c}{\textbf{REL}} \\
    \textbf{Language} & \textbf{Family} & \textbf{Region} & \textbf{Speakers}  & \textbf{Lang.} & \textbf{Source} & \textbf{Split Sizes} & \textbf{Source} & \textbf{Total Size} \\
    \midrule
    Amharic (\texttt{amh}) & Afro-Asiatic / Semitic & East & 33M & English & Global Voices & --/ 899/ 1037 & JW300 & 667K \\
    Hausa (\texttt{hau}) & Afro-Asiatic / Chadic &West& 63M & English & WMT2021: Khamenei.v1, Premium Times, Global Voices & 5865/ 1300/ 1500 & JW300 & 236K \\
    Kinyarwanda (\texttt{kin}) & NC / Bantu & East & 12M & English & Voice of America & --/ 460/ 1006 & JW300 & 485K \\
    Chichewa (\texttt{nya}) & NC / Bantu &East \& Central & 14M &English & Voice of America & --/ 483/ 1004 & JW300 & 775K \\
    chiShona (\texttt{sna}) & NC / Bantu &East \& Central & 12M & English & Voice of America & --/ 556/ 1005 & JW300 & 761K \\
    isiXhosa (\texttt{xho}) & NC / Volta-Niger &West& 9M & English & Voice of America & --/ 486/ 1002 & JW300 & 991K \\
    
    \bottomrule
  \end{tabular}
  }
  \vspace{-3mm}
  \caption{\textbf{Languages and Data Details for new languages added to the MAFAND-MT Corpus}. Language, family (NC: Niger-Congo), number of speakers, news source, news (\texttt{NEWS}), and religious domain (\texttt{REL}) data split. }
  \vspace{-4mm}
  \label{tab:data_stat_newer}
  \end{center}
\end{table*}

\begin{figure*}[t]
    \centering
    \includegraphics[width=0.69\linewidth]{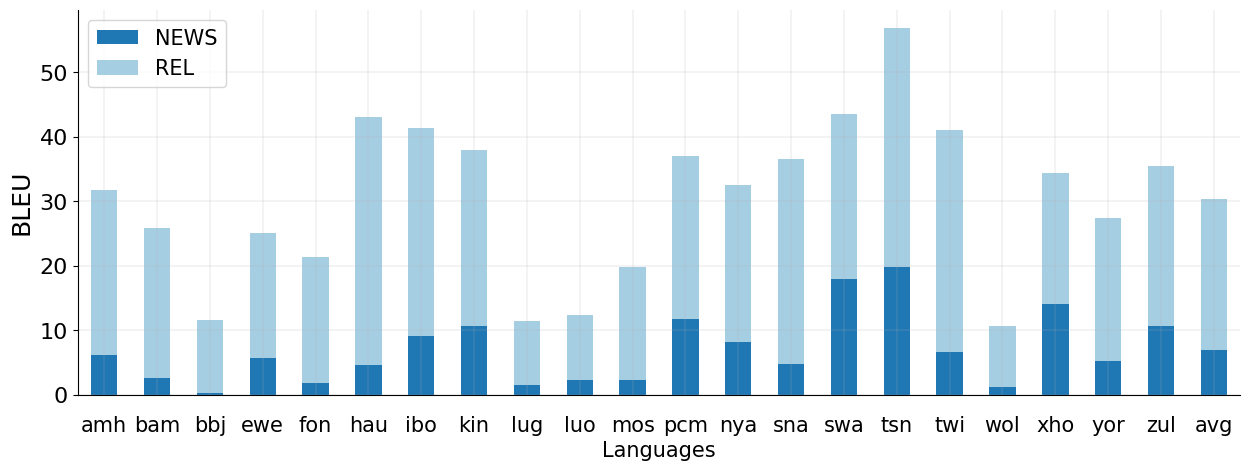}
    \includegraphics[width=0.69\linewidth]{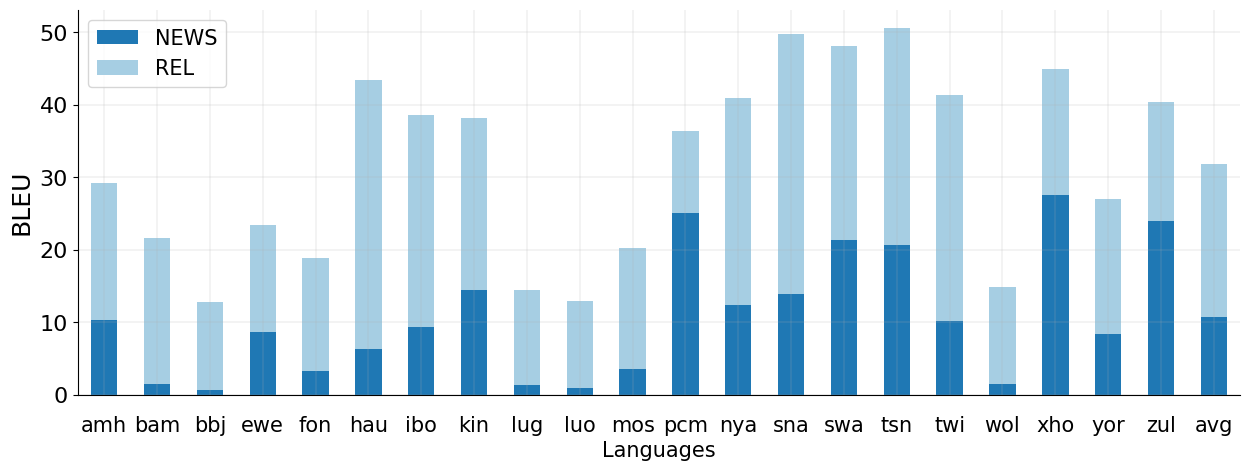}
    \caption{\textbf{Domain shift} of M2M-100 Transformer models trained on en/fr-xx (top) or xx-en/fr (bottom) \texttt{REL} domain and tested on the \texttt{NEWS} vs. \texttt{REL} domains.}
    \label{fig:domain_shift_newer}
    
\end{figure*}

\begin{table}[ht]
 \begin{center}
 \scalebox{0.70}{
  \begin{tabular}{p{13mm}p{8mm}rrrrrrrr}
    \toprule
     & \textbf{Tuned} & \textbf{hau} & \textbf{hau}\\
   \textbf{Evaluation} & \textbf{on} & \textbf{Khamenei } & \textbf{newer}\\
     \textbf{Domain} & \textbf{NEWS} & \textbf{dataset} & \textbf{dataset} \\
    \midrule
    \textit{en/fr-xx} \\ 
    FLORES & \xmark &$2.6$ & $2.6$    \\
    FLORES & \cmark &$4.0$ & $13.0$  \\
    REL & \xmark &$1.2$ & $1.2$   \\
    REL & \cmark &$3.7$ & $8.8$ \\
    \midrule
    \textit{xx-en/fr} \\ 
    FLORES & \xmark & $8.0$ & $8.0$ \\

    FLORES & \cmark & $16.3$ & $14.6$   \\
   REL & \xmark &$6.4$ & $6.4$  \\
    REL & \cmark &$3.8$ & $7.2$  \\
    \bottomrule
  \end{tabular}
 }
 \caption{\textbf{spBLEU on Wikipedia domain} (FLORES) and REL for M2M-100 before (\xmark) and after (\cmark) fine-tuning on \texttt{NEWS}. }
 \vspace{-3mm}
  \label{tab:dshift_flores_newer}
  \end{center}
\end{table}

\subsection{Evaluation dataset creation}
We translated about 1500 English sentences selected from the Voice of America (VOA) news platform to four more African languages: \textbf{Kinyarwanda (kin)}, \textbf{Shona (sna)}, \textbf{Chichewa (nya)}, and \textbf{IsiXhosa (xho)}. The 1500 sentences were divided into \texttt{DEV} and \texttt{TEST} split. Although the news articles are from VOA based in US, we ensured that the articles are related to events in Africa. Our choice of VOA is because it has an open license. We also added new evaluation dataset for \textbf{Amharic (amh)}, and increase the training data for \textbf{Hausa (hau)} by over 2K sentences. \autoref{tab:data_stat_newer} provides the data splits of the new evaluation data. We provide more details for Amharic and Hausa below. 

\paragraph{Amharic (amh)} We combined the Global Voices corpus\footnote{\url{https://opus.nlpl.eu/GlobalVoices.php}}  on OPUS~\citep{tiedemann-2012-parallel} with new articles from the Global Voices website\footnote{\url{https://am.globalvoices.org/}}. In total, we have 1,936 parallel sentences that we divide into \texttt{DEV} and \texttt{TEST} splits. 

\paragraph{Hausa (hau)} The Hausa Khamenei\footnote{\url{https://www.statmt.org/wmt21/translation-task.html}} corpus contains 5,898 sentences, we split them into \texttt{TRAIN} (3,098), \texttt{DEV} (1,300), and \texttt{TEST} split (1,500). We noticed that this dataset was created in Iran, which is not the geographical location of Hausa speakers. To diversify the texts, we decided to add 2767 \textit{newly translated} sentences from Global Voices and Premium times news websites which covers more Nigerian and West African news -- which is the location of native speakers of Hausa. In total, the training sentences increased to 5,865. 

\subsection{Additional experiments}
\paragraph{Domain shift} We extended the domain shift analysis on \autoref{fig:domain_shift} to the new languages. The results are quite similar. \autoref{fig:domain_shift_newer} shows the new result. 

\paragraph{Generalization of Hausa news corpus} In \autoref{tab:dshift_flores}, we show the generalization of the M2M-100 model trained on the \texttt{NEWS} domain to other domains like \texttt{REL} and \texttt{FLORES} (Wikipedia domain). We observe a poor generalization for the Hausa news corpus based on the Khamenei corpus to other domains. \autoref{tab:dshift_flores_newer} shows that by adding more contemporary news articles (2,767 sentences) from Premium times and Global Voices, we improved the spBLEU by large points especially in the EN-HAU direction ($4.0 \rightarrow 13.0$) for FLORES and ($3.7 \rightarrow 8.8$) for the REL domain (based on JW300). Although, we experienced slight drop in the \textit{xx-en/fr} direction for FLORES.

\end{document}